\documentclass[10pt, a4paper]{article}

\usepackage{graphicx}
\usepackage{subfig}
\usepackage{commath}
\usepackage{amsthm}
\usepackage{hyperref}
\usepackage{color}
\usepackage[table]{xcolor}
\usepackage{multirow}
\usepackage{booktabs}

\usepackage[english]{babel}
\usepackage{amssymb,amsmath,amsthm, amsfonts}

\RequirePackage[left=2cm,%
                right=2cm,%
                top=2.25cm,%
                bottom=2.25cm,%
                headheight=12pt,%
                letterpaper]{geometry}%

\theoremstyle{definition}

\title{Improving LBP and its variants using anisotropic diffusion}

\author{Mariane Barros Neiva$^{1,3}$, Patrick Guidotti$^{2}$,  Odemir Martinez Bruno$^{1,3}$}
\date{}
\begin{document}
\maketitle
\noindent{$^1$S\~{a}o Carlos Institute of Physics, University of S\~{a}o Paulo, S\~{a}o Carlos - SP, PO Box 369, 13560-970, Brazil.\\Scientific Computing Group - http://scg.ifsc.usp.br}

\noindent{$^2$Department of Mathematics, University of California Irvine, 340 Rowland Hall, Irvine, CA 92697, USA}

\noindent{$^3$Institute of Mathematics and Computer Science, University of S\~ao Paulo, USP, Avenida Trabalhador s\~ao-carlense, 400, 13566-590, S\~ao Carlos, SP, Brazil}

\begin{abstract}
The main purpose of this paper is to propose a new preprocessing step in order to improve local feature descriptors and texture classification. Preprocessing is implemented by using transformations which help highlight salient features that play a significant role in texture recognition. We evaluate and compare four different competing methods: three different anisotropic diffusion methods including the classical anisotropic Perona-Malik diffusion and two subsequent regularizations of it and the application of a Gaussian kernel, which is the classical multiscale approach in texture analysis. The combination of the transformed images and the original ones are analyzed. The results show that the use of the preprocessing step does lead to improved texture recognition. 
\end{abstract}

\section{Introduction}
\label{S:1}
Computer vision has became an important tool for industrial, scientific, and entertainment applications. The use of image and video processing in order to automate activities that used to require significant amount of time and effort is the main reason for this and justifies continued research towards better and better algorithms. An important task in vision is pattern recognition. It can be performed by using different characteristics of an image such as shape or color. Recently, the use of texture as an image feature proved to be a viable alternative \cite{kim1999statistical, haralick1979statistical, garcia2014local, costa2012efficient}. While a precise definition of texture is not agreed upon, humans naturally perceive it and, in fact, make use of it to recognize objects. It is therefore important to study its efficacy in image recognition.

Over the years, many scientists have described methods to extract features from texture. These can be divided into six main categories: statistical, stochastic, structural, spectral, complexity- and agent-based methods \cite{materka1998texture}. Each analyzes the image in a different way either by considering local patterns, Fourier spectrum, fractal dimension or other relevant quantities. However, more can be done. Usually, texture extraction methods are applied to the original dataset. If the image is not good and salient characteristics are not pronounced, texture analysis methods may not work as desired. 

In the latter case, one possibility is to enhance the image prior to attempting to classify it by texture. This paper aims at filling this gap by applying a preprocessing method to the original dataset and, subsequently performing texture extraction. Three databases will be tested with this approach: Brodatz, Vistex and UspTex. While, here, the technique is applied to classification, the approach can be utilized in combination with any other task that requires texture features, such as segmentation, synteshis, detection and prediction. 

The core of the proposed method is the application of a preprocessing step. Different methods of diffusion were chosen for this task to check whether multiscale technique improves feature extraction. The multiscale nature of vision is implicit in our visual system in the cortex area where receptive fields separate bars, corners, lines and edges of an image \cite{koenderink1987representation}. Usually, multiscale resolution is achieved by a Gaussian kernel with different variances (i.e. values of $\sigma$). Using a Gaussian filter, however, alters edges, which are typically a very important features of an image. Gaussian filtering amounts to applying isotropic diffusion which is insensitive to direction and hence to geometry. This problem can be overcome by the use of nonlinear anisotropic diffusion. Edges are implicitly located by the method, which then prevents their blurring, effectively confining the effect of diffusion within areas separated by the detected edges.

Beside isotropic diffusion, the paper also uses three different anisotropic diffusion methods: Perona et. al. anisotropic diffusion \cite{perona1994anisotropic}, and two regularizations of it proposed by Guidotti et al.: a forward-backward regularization of Perona-Malik \cite{guidotti2012backward} and another using fractional derivatives (NL) \cite{guidotti2009new}. Experiments are performed with all methods and comparisons are presented. Without any prior hypothesis on whether isotropic or anisotropic methods are the best, the goal is to determine what preprocessing method enhances which characteristics analyzed by feature extraction.

Also, the paper has the goal to improve one of the most used descriptors nowadays. The Local Binary Pattern method, proposed in 2002 by Ojala et. al. \cite{ojala2002multiresolution} has become well-known as a descriptor in computer vision due its effectiveness and simplicity. Over the years, different extensions of this method were developed to improve the descriptor and it is still very much used in practical applications. Therefore, the paper uses these local pattern extractors as descriptors and tries to enhance them by using the preprocessed images. The paper is structured as follows: Section \ref{ref:background} presents the texture description methods used: LBP, CLBP, LBPV, LBPFH, LTP and CSLBP as well as briefly introduces the four diffusion methods. Section \ref{ref:proposal} details the proposed approach combining preprocessing with the descriptor methods along with a classifier. Section \ref{ref:results} shows the results for texture classification using four datasets and, finally, Section \ref{ref:conclusion} draws the main conclusions of our work.

\section{Background}\label{ref:background}

Different feature extraction methods will be used in order to test the possible advantages of applying diffusion methods prior to feature extraction. These descriptors along with the four preprocessing methods are described in this section.

\subsection{Feature Extraction Methods}\label{sec:fem}
\subsubsection{Local Binary Pattern}

This statistical method \cite{ojala2002multiresolution} extracts features analyzing the relationship between pixels and their neighbours in search of local patterns. Values are obtained from the threshold of the central pixel in relation to its $P$ neighbours (Equation \ref{eqLBP}).  In Equation \ref{eqLBP}, $g_c$ is the value of the central pixel and $g_p$ are the values of the neighbours according to a radius $r$ around the main pixel. The algorithm ends when all pixels are used as threshold (i.e. as central pixel) generating $M\times N$ numbers corresponding to the size of the image. A histogram is constructed with patterns according to Equation \ref{eqH}.

\begin{equation}
LBP_{P,r} = \sum_{p=0}^{P-1} s( g_p - g_c)\,2^p,\: s(x) =\begin{cases}
1,&x \geq 0, \\ 0,& x < 0,
\end{cases}
\label{eqLBP}
\end{equation}

\begin{gather}\label{eqH}
H(k) = \sum_{i=1}^{M} \sum_{j=1}^{N} \text{ bool}\bigl( LBP_{P,r}(i,j) ,k\bigr),\\
\text{ where }  k \in [0,K]\text{ and}\\
\text{ bool}(x,y) = \begin{cases}
 1,& x = y \\ 0,& \text{otherwise}
\end{cases}
\end{gather}
The basic approach uses a $3\times 3$ window, i.e. P = 8, r = 1. Therefore, the histogram, and, consequently, the feature vector, will have $2^8 = 256$ entries.

\subsubsection{Local Binary Pattern Variance}
To provide rotation invariant features, Ojala et. al. \cite{ojala2002multiresolution} proposed a method to combine into a single pattern all basic LBP patterns that only differ by a rotation by counting the number of bits that change from 0 to 1 or from 1 to 0. These patterns are computed based on a metric U that checks uniformity:

\begin{equation}
 U(LBP_{P,R}) = \big |s(g_{p-1} -g_c) - (g_0 - g_c)\big | + 
 \sum_{p = 1}^{p = P-1} \big |s(g_{p} -g_c) - (g_{p-1} - g_c)\big |.
\end{equation}
Therefore, uniform rotation invariant LBP is computed by:
\begin{equation}
  LBP_{P,R}^{riu2} (i,j) = \begin{cases} 
   \sum_{p=0}^{P - 1} s(g_p - g_c),&\text{if } U(LBP_{P,R} (i,j)) < 2 \\
   P + 1,&\text{otherwise}\end{cases}.
\end{equation}
Then, in order to add contrast information to $LBP_{P,R}^{riu2}$, Guo et.al. \cite{guo2010completed} proposed the use of variance in combination with the uniform rotation invariant descriptor. It is computed by
\begin{equation}
  VAR_{P,R} = \frac{1}{P} \sum_{p = 0}^{P - 1}\bigl[
  g_p - (\frac{1}{P}\sum_{p = 0}^{P - 1} g_p)\bigr]^2
\end{equation}
In practice, variance is used to compute the strength of each pattern $k\in[0,K]$ in the $LBP_{P,R}^{riu2}$ of an image $I$ of size $M\times N$:
\begin{gather}
  LBPV_{P,R} (k) = \sum_{i=1}^M \sum_{j=1}^N \omega\bigl( LBP_{P,R}^{riu2} (i,j),k\bigr),\:
  k \in [0,K] \\
  \omega\bigl( LBP_{P,R}^{riu2},k\bigr) = \begin{cases} 
   VAR_{P,R} (i,j),&\text{if } LBP_{P,R}^{riu2} = k,\\
   0,&\text{otherwise}\end{cases},
\end{gather}
finally, LBPV is used as feature vector for the image $I$. 

\subsection{Complete Local Binary Pattern}

One of the best extensions of classic LBP nowadays is the Complete Local Binary Pattern descriptor. Proposed by Guo et. al. \cite{guo2010completed}, the technique improves the inspirational method by adding information of sign and magnitude of the local patterns. 

The sign and magnitude are computed from matrix  LSDMT where each pixel $p$ is calculated as: 

\begin{equation}
LSDMT_p = g_c - g_p. \label{eq:ldsmt}
\end{equation}
where $g_c$ and $g_p$ are intensity value of central pixel of current evaluated window and and the value of the pixel at position $p$. From LSDMT two matrices are computed: CLBP\_S and CLBP\_M. The first is computed by the sign of the original matrix while the second is formed by the absolute values of it. 

Finally, three different information are used to   generate the features: the image itself (CLBP\_C), CLBP\_S (which represents the same information contained in LBP) and CLBP\_M.  CLBP\_M  is a a matrix  of decimals  must be converted by:

\begin{equation}
CLBP\_M_{P,r}=\sum_{p=0}^{P-1}t(g_p,g_c)2^{p},\quad t(x,c)=\begin{cases}1,&x\geq c\cr 0,&x<c \end{cases}
\end{equation}
 
A histogram of the concatenation of the three matrices above is used as feature vector in this method. Results shows that the descriptor output high recognition rates with the addition of sign and magnitude information of local patterns. Like LBP, P and r are used as 8 and 1 respectively in the experiments.

\subsection{Local Binary Pattern Histogram Fourier Features}

This method had the advantage to be not only rotation invariant locally but also globally \cite{ahonen2009rotation}. It first computes the non-invariant LBP histograms and from them, the method calculates the global rotation invariant features from the basic approach. 
Likely the classical LBP, the method uses a 3x3 window around the central pixel to extract binary features and reduces the histogram applying a rule that if pattern only differs by rotation, it is computed to the same bin. Therefore, patterns such as 110 and 011 are considered the same in the method. Then, the paper states that if original image is rotated by a certain degree $\alpha$  ($\alpha = a\frac{360}{P}$) where P is the number of pixels evaluated in each window, the histogram is also cyclic shifted. To reach global invariance, the Discrete Fourier Transform is computed according to equation:

\begin{equation}
H(n,u) = \sum_{r = 0}^{P-1} h(U_P(n,r))e^{-i2\pi u r/P}.
\end{equation}
where $r$ is the rotation of the pattern and $U_P$ is the pattern obtained from rotation invariant. The original article shows that features in spectrum domain are invariant to cyclic features and therefore rotation invariant.

\subsection{Local Ternary Pattern}

In this forth descriptor, instead of representing the pattern about a central pixel with only two bits, three different values are used to represent a pattern \cite{tan2010enhanced}. In order to extract features around a certain value $c$ with a small deviation $k$ and neighbor pixel value $p$, the result of the threshold is as follows:

\begin{equation}
\begin{cases}
1, &\text{if } p > c + k,\\ 0, &\text{if }  p > c - k  \text{ and } p < c + k, 
\\ -1, &\text{if } p < c - k. 
\end{cases}
\end{equation}

If one computes the corresponding histogram, one would be able to perceive that the number of features is very high. Therefore, in order to minimize the final vector size, two binary patterns are extracted from the ternary ones. Finally, an histogram is obtained and used as feature vector. 

\subsection{Center-symmetric Local Binary Pattern}

This additional local feature descriptor combines the power of the well-known LBP and the famous SIFT \cite{heikkila2006description}. The method first applies a noise removal in each analyzed window. The filter is able to preserve edges while removing noise which then allows LBP to work better as it relies on edge information. 

After filtering a pattern is extracted for each pixel in the image similarly to what occurs in classic LBP. However, in this case, features are extracted comparing the center-symmetric pairs of pixels meaning the pixels that are in the opposite position according to the central pixel. Thus, while LBP uses a 3x3 window to produce 256 features, CSLBP computes a feature vector of only 16 values. In addition, in this method, the threshold used is set to a value T, in the experiments T = 0.01, to improve results in flat image regions:

\begin{align}
 CSLBP(x,y) &= \sum _{i=0}^{P/2 - 2} s(n_i - n_{i+(P+2)})2ˆi,\\ s(x) 
 &= \begin{cases}
1, &x > T,\\ 0, &\text{ otherwise,}
\end{cases}
\end{align}
where P is the number of neighbors considered and $n_i$ and $n_i + (P/2)$ are the symmetric pairs. In the experiments a grid of 4x4 is used obtaining a final histogram of 256 features.

\subsection{Diffusion Methods}

The key methods implemented in this paper are of diffusion type. Linear and nonlinear, isotropic and anisotropic diffusions are used to create a new set of transformed images at different scales. A brief description of each of them follows.

\subsubsection{Isotropic Diffusion}

The simplest and oldest form of multiscale resolution is obtained by applying a Gaussian kernel to the texture with varying smoothing parameter $\sigma$. A 2D Gaussian filter is calculated by convolution with
\begin{equation}
  G(x,y) = \frac{1}{{2\pi \sigma^2}} e^{-\frac{x^2 + y^2}{2 \sigma^2}} \label{eq:gaussian}
\end{equation}
For such an isotropic diffusion, smoothing is applied equally in all regions of the image and in all directions. Consequently, information from edges and from smooth regions gets mixed up as the scale parameter $\sigma$ is increased. High frequency areas such as edges and corners are, however, an important feature of texture and this information is lost in the multiscale pyramid obtained by this filtering procedure.

\subsubsection{Perona-Malik's Anisotropic Diffusion}

Perona et. al. \cite{perona1994anisotropic} were among the first to propose and to develop non-linear anisotropic diffusion methods. Until then a multiscale resolution was obtained by convolving the image intensity function with a family of Gaussian filters with different variances ($\sigma$ values) leading to a sequence of images of varying levels of detail as explained above. The higher the $\sigma$, the coarser the details. Convolution with a Gaussian is given by
\begin{equation}\label{eq:gaussianc}
  I(x,y,t) = I_o(x,y) \ast G(x,y;t)
\end{equation}
for G as Equation \ref{eq:gaussian} and can be thought of representing the solution of the heat equation
\begin{equation}
I_t = \nabla^2 I = I_{xx} + I_{yy}
\end{equation}
with initial datum $I_0(x,y)$ and where $t=\sigma^2$.

As mentioned before, using a Gaussian kernel to generate different levels of details, however, leads to the disappearance of edges, 
which are a source of important information on an image. 

A multiscale pyramid can be visualized by means of a tree, where, at one level, the shape of the tree can be
recognized, and, at next level, branches and leaves are added. Eventually, increasing the resolution level, all 
details including leaves' veins are present. Regardless of resolution, edges are present and well defined and, as 
such, should be preserved for visual tasks. 

\begin{figure}%
    \centering
    \subfloat[Original Image]{{\includegraphics[scale=0.85]{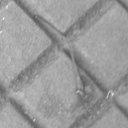} }}%
    \qquad
    \subfloat[Gaussian Diffusion - $\sigma$ = 2.0]{{\includegraphics[scale=0.85]{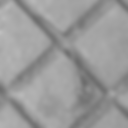} }}%
    \qquad
    \subfloat[Perona-Malik Diffusion - 15th iteration]{{\includegraphics[scale=0.85]{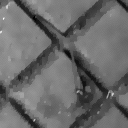} }}%
    \caption{Comparison between classic Gaussian diffusion and classic non linear diffusion. Adapted from: \cite{usptex}}%
    \label{fig:peronacomparison}%
\end{figure}

This was one of the reasons which motivated Perona et. al. \cite{perona1994anisotropic} to propose the anisotropic diffusion 
given by
\begin{equation}\label{eq:aniso}
 I_t = \operatorname{div}\bigl( c(x,y,t)\nabla I\bigr) = c(x,y,t) 
 \Delta I + \nabla c \cdot \nabla I 
\end{equation}
The goal is then to find a function $c$ that would result in the output image satisfying three main 
requirements
\begin{itemize}
\item[] {\bf Causality}. No fake details should be generated in coarser resolutions. 
\item[] {\bf Immediate Location}. At each resolution, edges should be sharp and semantically meaningful to the actual resolution.
\item[] {\bf Piecewise Smoothing}. Blurring should privilege intra-regions over inter-regions. 
\end{itemize}
If $c$ is constant, $\nabla c$ vanishes and the diffusion is linear and isotropic. As intra-region blurring is 
to be privileged over inter-region blurring, $c$ should have a lower value at inter-region boundaries. 
Thus, in \cite{perona1994anisotropic}, $c$ is related to the gradient of the image itself, which is used 
as an edge detector
\begin{equation}
c(x,y,t) = g\bigl( |\nabla I (x,y,t)|\bigr).
\end{equation}
The function $g$ is taken to be related to the inverse of the gradient square and regulates the amount of diffusion applied in the region. Since, at inter-region boundaries, the gradient is large, blurring is lower. One of the functions proposed in \cite{perona1994anisotropic} is given by
\begin{equation}\label{eq:g1}
g(\nabla I) =  \frac{1}{1+ (\frac{|| \nabla I ||}{\kappa})^2}
\end{equation}
and will be used in the sequel. The parameter $\kappa$ is user-defined and controls the threshold value beyond which
sharp transitions are considered edges and are hence preserved. According to the authors, the chosen form of $g$
privileges large regions. Image \ref{fig:peronacomparison} compares linear diffusion to the nonlinear,
anisotropic diffusion of \cite{perona1994anisotropic}. It is clear that edges are better maintained by the latter.
While fake details, such as staircasing, are produced by this type of anisotropic diffusion, blur is effectively
confined to regions between edges, delivering on at least two of the three properties above.

\subsubsection{Forward-Backward Regularization Diffusion}\label{sec:fbr}

In \cite{guidotti2013image} it is noted that, while Perona et. al. \cite{perona1994anisotropic} method is an 
interesting and effective model, it is not without shortcomings due to the generation of artificial edges, a
phenomenon known as staircasing. Therefore, the authors propose a regularization characterized by two parameters
$p\in(1,\infty)$ and $\delta>0$ with the aim of avoiding staircasing without completely discarding the edge
sharpening power of the original Perona-Malik model. The equation proposed reads as follows

\begin{equation}\label{eq:patrick1}
 I_t=\nabla\cdot\Big(\big[\frac{1}{1 + K^2|\nabla I|^2} + 
 \delta |\nabla I|^{p-2}\big]\nabla I \Big),
\end{equation}

According to the experiments in \cite{guidotti2013image}, best results are obtained for a parameter $p$ which is 
close (but not equal) to 1. The addition of this two parameters allows the control of desired gradient growth and 
also effectively bounds it to a maximal size. This is due to the forward-backward nature of the equations and the 
fact that the backward regime is confined to $\bigl[ 1 < |\nabla I|  < M(\delta,p)\bigr]$. As a consequence, 
staircasting is replaced by (micro)-ramping, the steepness of which is controlled by the values of $p$ and 
$\delta$. The right choice of the parameters is essential. For $p=1$, for instance, the backward regime is 
unbounded, leading to staircasing. An experiment using $p=2$ is shown in \cite{guidotti2012backward}. \cite{guidotti2013image} shows good results when employing Equation \ref{eq:patrick1} for denoising and 
debluring and it will be tested here as a feature enhancer for texture classification. 

\begin{figure}%
    \centering
    \subfloat[Gaussian Diffusion - $\sigma$ = 2.0]{{\includegraphics[scale=0.85]{gausian2.png} }}%
    \qquad
    \subfloat[Perona-Malik Diffusion - 15th iteration]{{\includegraphics[scale=0.85]{perona15.png} }}%
      \qquad 
    \subfloat[Forward-Backward Regularization Diffusion - 15th iteration]{{\includegraphics[scale=0.85]{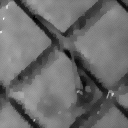} }}%
      \caption{Comparison between classic Gaussian diffusion, classic non linear diffusion and a regularization of perona-malik equation. Adapted from: \cite{usptex}}%
    \label{fig:fbr}%
\end{figure}

Figure \ref{fig:fbr} shows the comparison between the three approaches explained so far. The visual difference 
between Gaussian and anisotropic methods is very evident. However, comparing both anisotropic methods, it 
is hard to see the micro-ramping versus staircasing while it does exist as can be seen in 
\cite{guidotti2013image}. 

\subsubsection{Nonlocal Anisotropic Diffusion}
In \cite{guidotti2009new} another approach is taken to regularize Perona-Malik and thus effectively avoid
staircasing without reducing Perona-Malik intrinsic sharpening feature. The trick is performed by the use a 
fractional derivatives in the edge-detector

\begin{equation}\label{eq:patrick2}
  I_t = \nabla (\frac{1}{1+ K^2|\nabla^{1-\varepsilon}I|^2}\nabla I) 
  \text{, where }\varepsilon \in (0,1)
\end{equation}

The main advantages of Equation \ref{eq:patrick2} are its mathematical well-posendness, on the one hand, and its 
simultaneous tendency to promote intra-region smoothing while solidly preserving edges for a long time. Latter is 
due to the fact that, as stated in Guidotti et. al. \cite{guidotti2009new}, ``...piecewise constant functions or 
characteristic functions of smooth sets in higher dimension can be shown to be equilibria for the evolution.'' 

If $\varepsilon = 0$, the equation reduces to Perona-Malik. However, positive values of $\varepsilon$ require the
calculation of fractional derivative. This can be done by Fourier transform considering periodic boundaries as follows:

\begin{equation}\label{eq:fractional}
  |\nabla^{1-\varepsilon}I| = \mathcal{F}^{-1}\Bigl(\operatorname{diag}
  \bigl[2\pi|k|^{-\varepsilon}\bigr] \mathcal{F}(|\nabla I|)\Bigr),
\end{equation}
where
\begin{equation}
  \mathcal{F}(I)(k) = \int_{\Omega} e^{-2\pi i k\cdot x} I(x)\, dx,
  \: k \in \mathbb{Z}^2.
\end{equation}
and $\operatorname{diag}\bigl[(m_k)_{k\in\mathbb{Z}^2}\bigr]$ denotes matrix multiplication with diagonal entries given by the sequence $(m_k)_{k\in\mathbb{Z}^2}$. 

This approach is capable to partially capture nonlocal information due the use of the ``wider support'' fractional
derivatives and thus enhancing edge detection robustness in the presence of noise.  This well-posed anisotropic
diffusion can be effectively used for denoising purposes due to its ability to differentiate between high frequency
global features (that really belong to the image) and local ones (noise). Further explanations and proofs are found in \cite{guidotti2009new}. 

\begin{figure}%
    \centering
    \subfloat[Gaussian Diffusion - $\sigma$ = 2.0]{{\includegraphics[scale=0.85]{gausian2.png} }}%
    \qquad
    \subfloat[Perona-Malik Diffusion - 15th iteration]{{\includegraphics[scale=0.85]{perona15.png} }}%
  \qquad   
      \subfloat[NonLocal Anisotropic Diffusion  - 15th iteration]{{\includegraphics[scale=0.85]{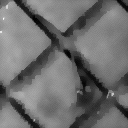} }}%
      \caption{Comparison between classic Gaussian diffusion, classic non linear diffusion and the non local diffusion. Adapted from: \cite{usptex}}%
    \label{fig:nonlocal}%
\end{figure}

Figure \ref{fig:nonlocal} shows results very comparable to what Figure \ref{fig:fbr} demonstrates where differences
between both anisotropic methods are very subtle while compared to Gaussian diffusion the difference is high in 
terms of edge preservation. 
 
\section{Proposed Method}\label{ref:proposal}
While typically used for deblurring and denoising purposes, Gaussian filters, Perona-Malik and the enhanced counterparts
\eqref{eq:patrick1} and \eqref{eq:patrick2} can also be used as a basic tool to obtain multiresolutions of an image. 
For these algorithms, in fact, coarser and coarser images are created over time. 
We will use these images to extract features.

Motivated by human vision where dots, corners and lines are separated in the cortex for object recognition
\cite{julesz1975experiments}, a multiscale resolution that can distinguish different features of the image and use
them separately can deliver a better and easier input to feature extraction methods. At different iterations (denoted
by the parameter $it$), different features appear, and the one that better represents the dataset will 
hopefully give a higher texture classification rate. 

The method works as follows: the original dataset is processed with the diffusion methods 
generating 150 new images for each one in the dataset. Then, each one of the images at one scale along the original is feature-extracted and one hit rate is computed.

\begin{figure}%
  \centering
  \includegraphics[width=14cm]{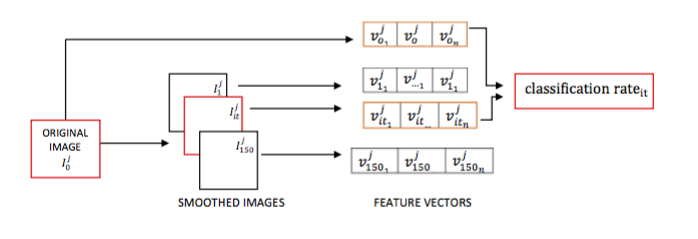}         
    \caption{Diagram of the proposed method. Images are first transformed by the preprocessing and then represented as
    characteristics extracted by texture analysis methods. The next step is to concatenate features of original
    images and analyzes scale. Finally, classification is performed.}\label{fig:diagram}
\end{figure}

Anisotropic methods are believed to have the advantage over isotropic ones to smooth while keeping important 
structures of the images such as edges and T-junctions. This advantage will be tested for feature extraction. 
Beyond just adding a preprocessing method to feature extraction, this paper has also the goal to compare 
the four different methods described above and analyze which one most enhances each type of texture features. 
An diagram of the approach is presented in Image \ref{fig:diagram}. 

As a case study, six feature extraction methods will be used to analyze characteristics of the images. These were chosen mainly because of the importance of the local binary pattern descriptor in the literature. The good results of the simple technique awoke the interest of researchers and lead to attempts at extending the method. This paper also offers a different approach in order to improve the classical one as well assome extensions of it. For the recognition step, K-Nearest Neighbor and Naive Bayes will be applied as a simple yet powerful classification model \cite{amancio2014systematic}. 

\section{Experimental Results}\label{ref:results}

\subsection{Datasets}

Three texture datasets are used to evaluate the approach: 

\begin{itemize}
\item \textbf{Brodatz Dataset \cite{B66}}: a well known, widely used dataset to test texture related algorithms. The set
contains 1110 images from 111 different classes. Images are grayscale and of size 200x200. Brodatz is
characterized by a lack of ilumination, scale, and viewpoint variation. As a well-known database of textures 
it should be tested anyway.
\item \textbf{Vistex Dataset \cite{Vistex}}: The Vistex dataset contains a total of 864 RGB color images, 16 in each class. 
It contains variations in illumination conditions and scale. Images are of size 128x128 and they are converted 
to grayscale images in the experiments below. 
\item \textbf{UspTex Dataset\cite{BackesB10}}: The dataset contains 12 color images per class for a total of 2292 images of 
size 128x128. Each class set is generated by taking non ovelarping windows of an original 512x384 image. All
images are converted to grayscale prior to preprocessing. 
\end{itemize}

\subsection{Performance Evaluation}

For simplicity and efficiency, KNN (k = 1) and Naive Bayes are used. A cross validation (10-fold) was 
performed to get a more reliable classification rate. For each dataset, preprocessing methods were applied
generating 150 scales.

For isotropic diffusion, $\sigma$ started as 0.5 and was incremented by 0.5 at each new iteration. 
For the first anisotropic method, implementation was done following \cite{perona1994anisotropic} with $g$ 
as in \eqref{eq:g1} with K = 1. For forward-backward regularization (Section \ref{sec:fbr}), parameters 
were chosen to be $\delta = 0.1 $ and $p = 1.1 $. Finally, for nonlocal anisotropic diffusion \ref{sec:fbr}, $\varepsilon$ was set to 0.1. The time derivation for convolution was set to 0.25 for all methods. Further experiments could be done evaluating the influence of the parametrization set. 

Each scale was evaluated separately to analyze which one better improved feature extraction (Image \ref{fig:diagram}). We joined the original image feature vector and the smoothed image vector from iteration $it$ to use as input for the KNN classification space.  As datasets have different characteristics, each method works in a different way and the $it$ scale that yields the best improvement in feature extraction is different. 

Analyzing the results for each dataset, the easiest, Brodatz, showed the best performance when Forward Backward Regularization (FBR) was combined with the CLBP descriptor. It can be noticed that any combination shown in Tables \ref{table:knn} and \ref{table:naivebayes} improves the classification rate when compared to the traditional approach. The most benefited by the proposal is CSLBP with a gain of 5.94\%. In general, FBR was the method that
most improved the classification rate for this dataset for both classifications. 

The second dataset, Vistex, produces the best classification rate by a combination FBR + CLBP (KNN, k = 1). Likely Brodatz,
images are better characterized when FBR is applied even though all combinations result in enhanced texture classification. However, the method with highest advantage in this case is LBPHF with an improvement of up to 11.23\%. 

Finally, for Usptex, the improvement in texture recognition obtained with the addition images preprocessed by FBR method was 6.33\% for the same CLBP method. This set was the most benefited by the addition of transformed images, the difference between classical approach and proposed combinations output the highest gains. The average improvement was of 10.25\% with lower gain around 6\% considering all descriptors tested in this experiment. 

\begin{figure}%
 \centering
  \subfloat[Brodatz]{{\includegraphics[width=0.33\textwidth]{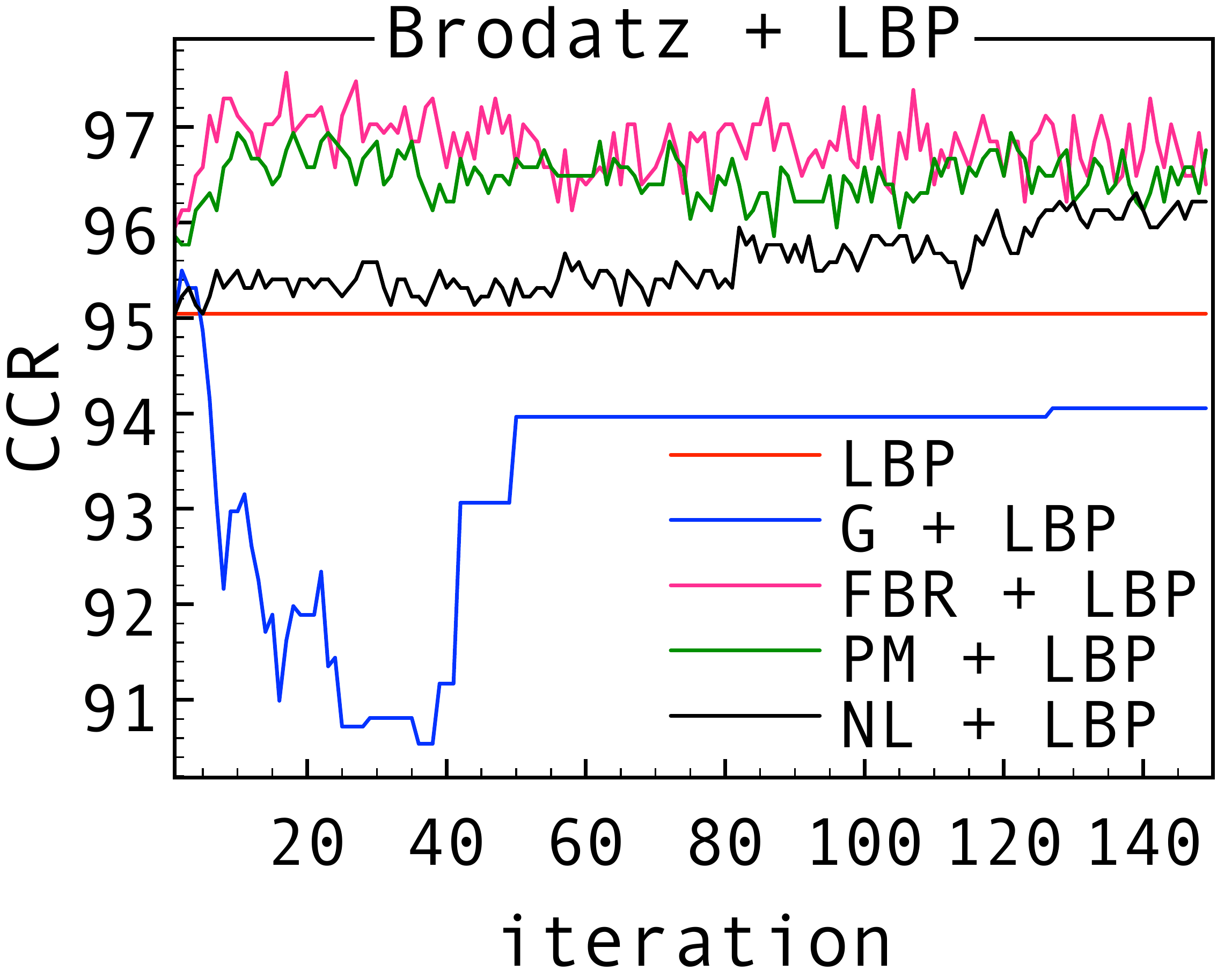} }}\subfloat[Usptex]{{\includegraphics[width=0.33\textwidth]{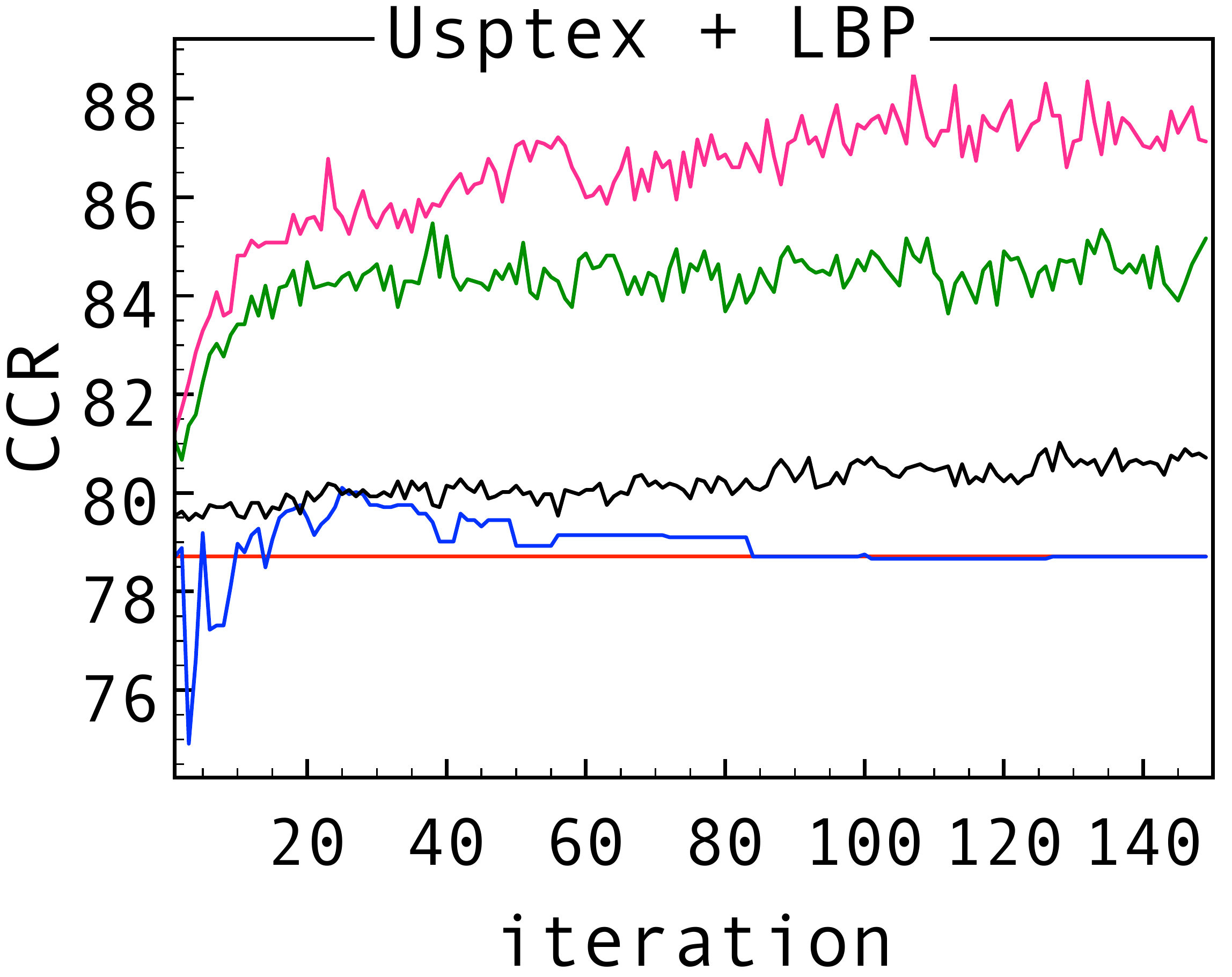} }}  \subfloat[Vistex]{{\includegraphics[width=0.33\textwidth]{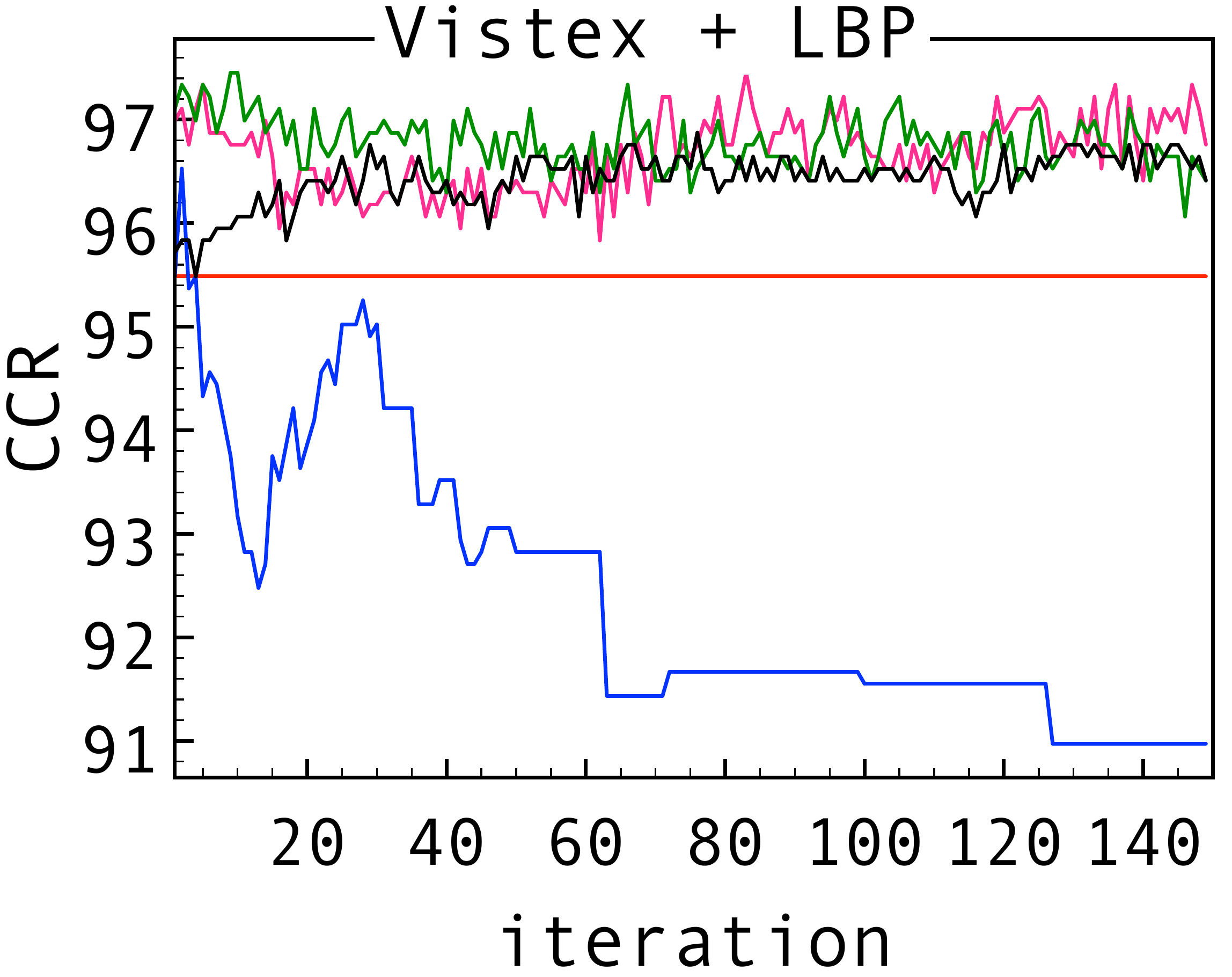} }}%
  \caption{Results of each iteration when preprocessing methods are combined with LBP (Naive Bayes) }  \label{plot:lbp}
\end{figure}

\begin{figure}%
 \centering
 \subfloat[Brodatz]{{\includegraphics[width=0.33\textwidth]{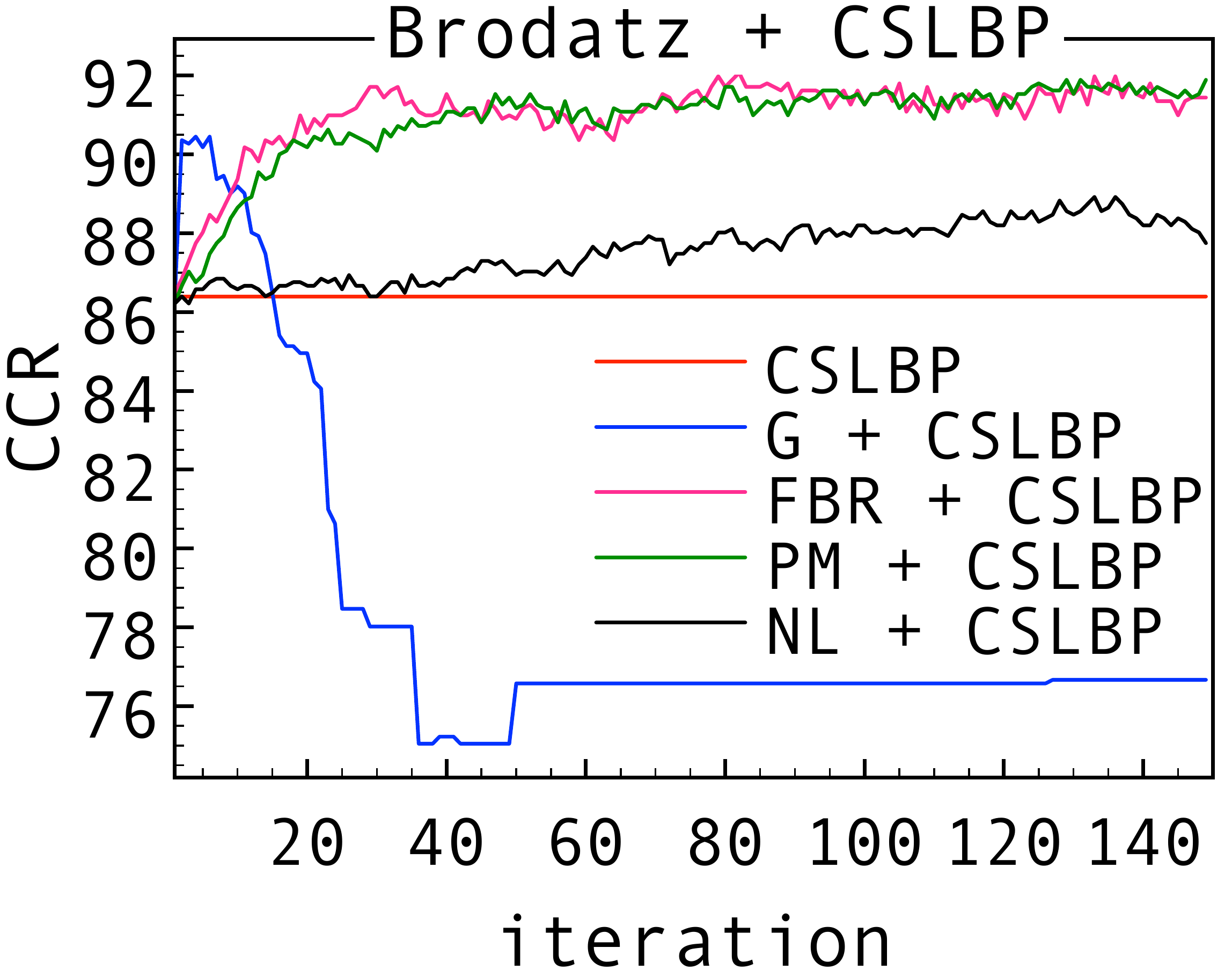} }}\subfloat[Usptex]{{\includegraphics[width=0.33\textwidth]{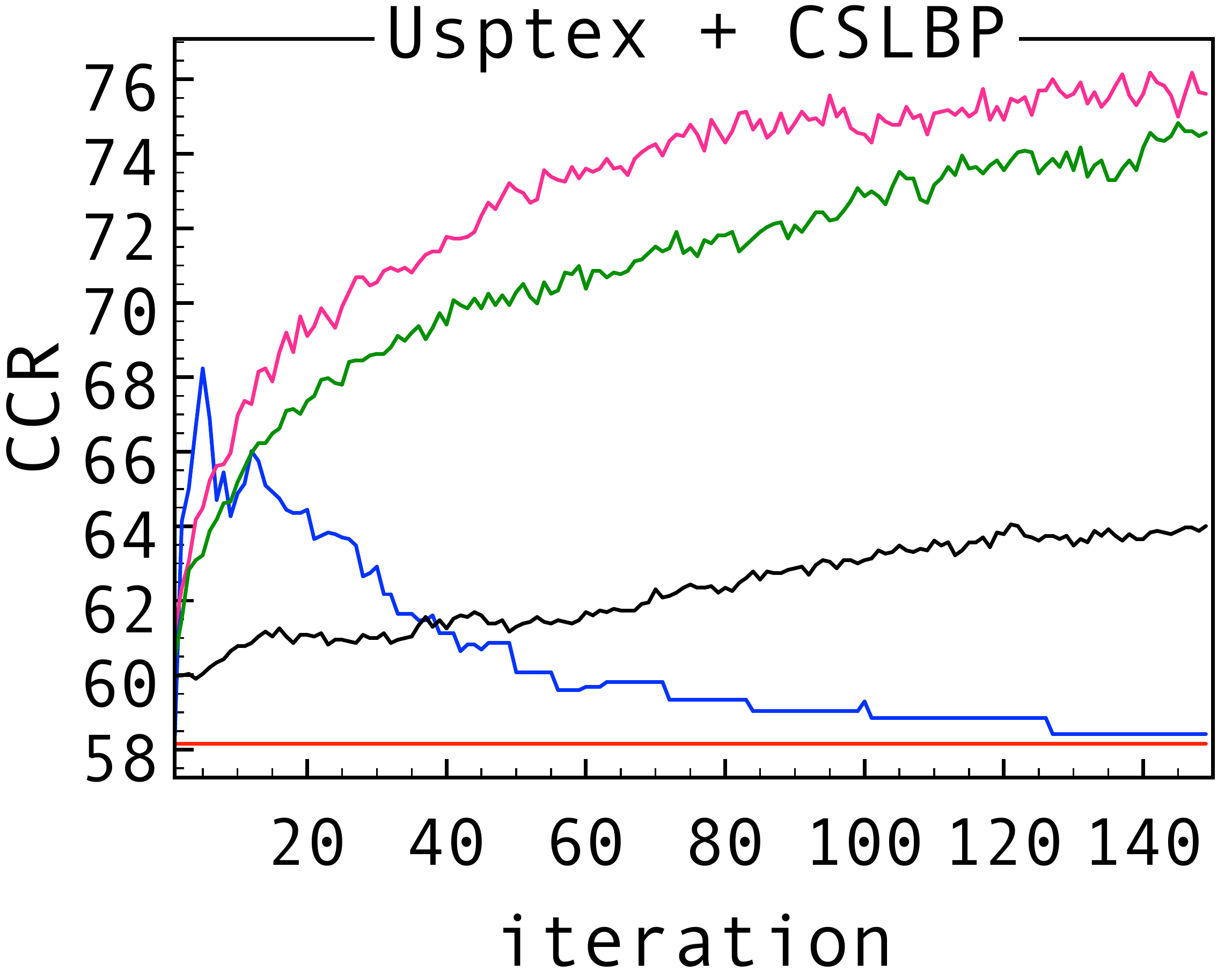} }}\subfloat[Vistex]{{\includegraphics[width=0.33\textwidth]{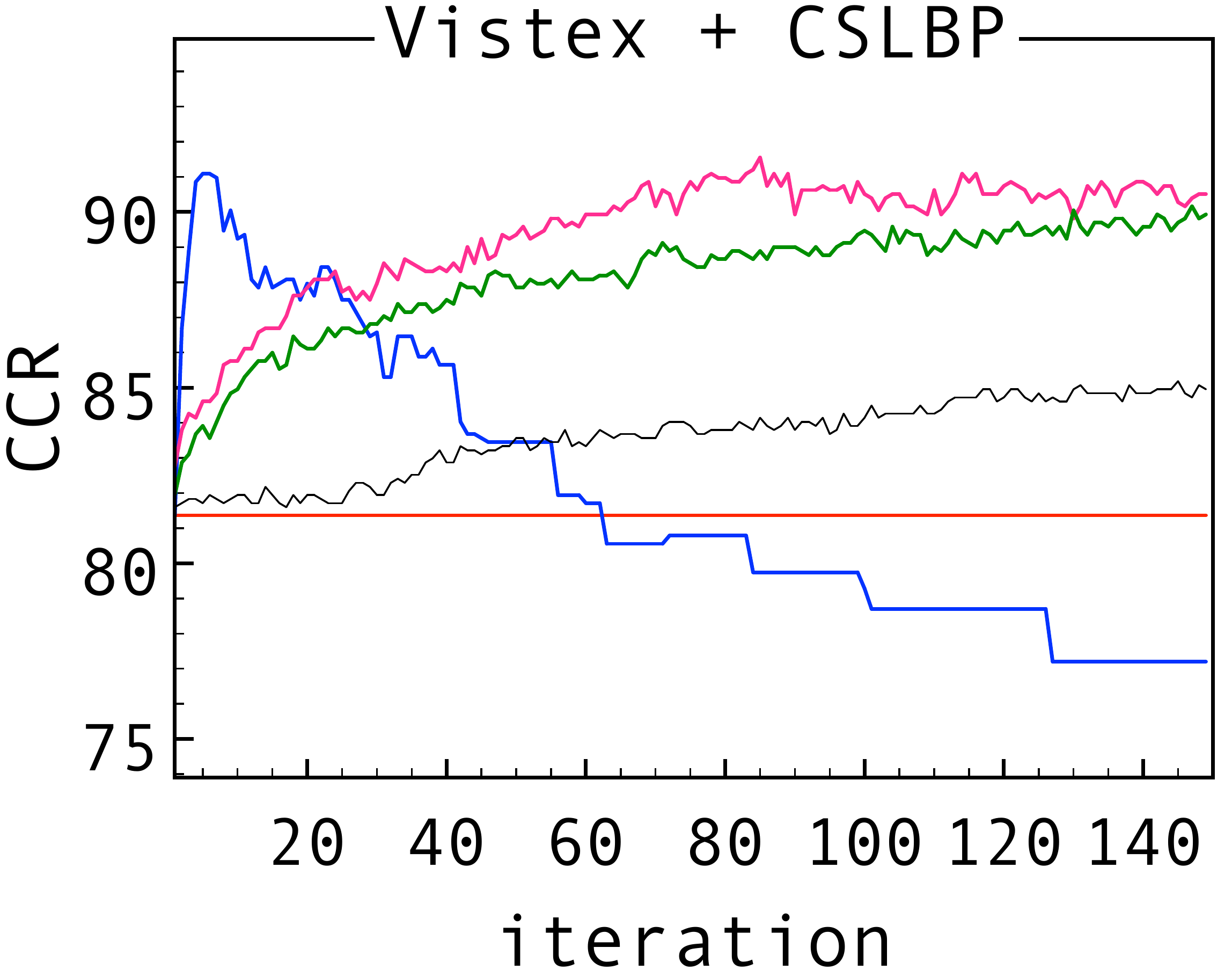} }}%
  \caption{Results of each iteration when preprocessing methods are combined with CSLBP (Naive Bayes) }\label{plot:cslbp}
\end{figure}

\begin{figure}%
 \centering
 \subfloat[]{{\includegraphics[width=0.33\textwidth]{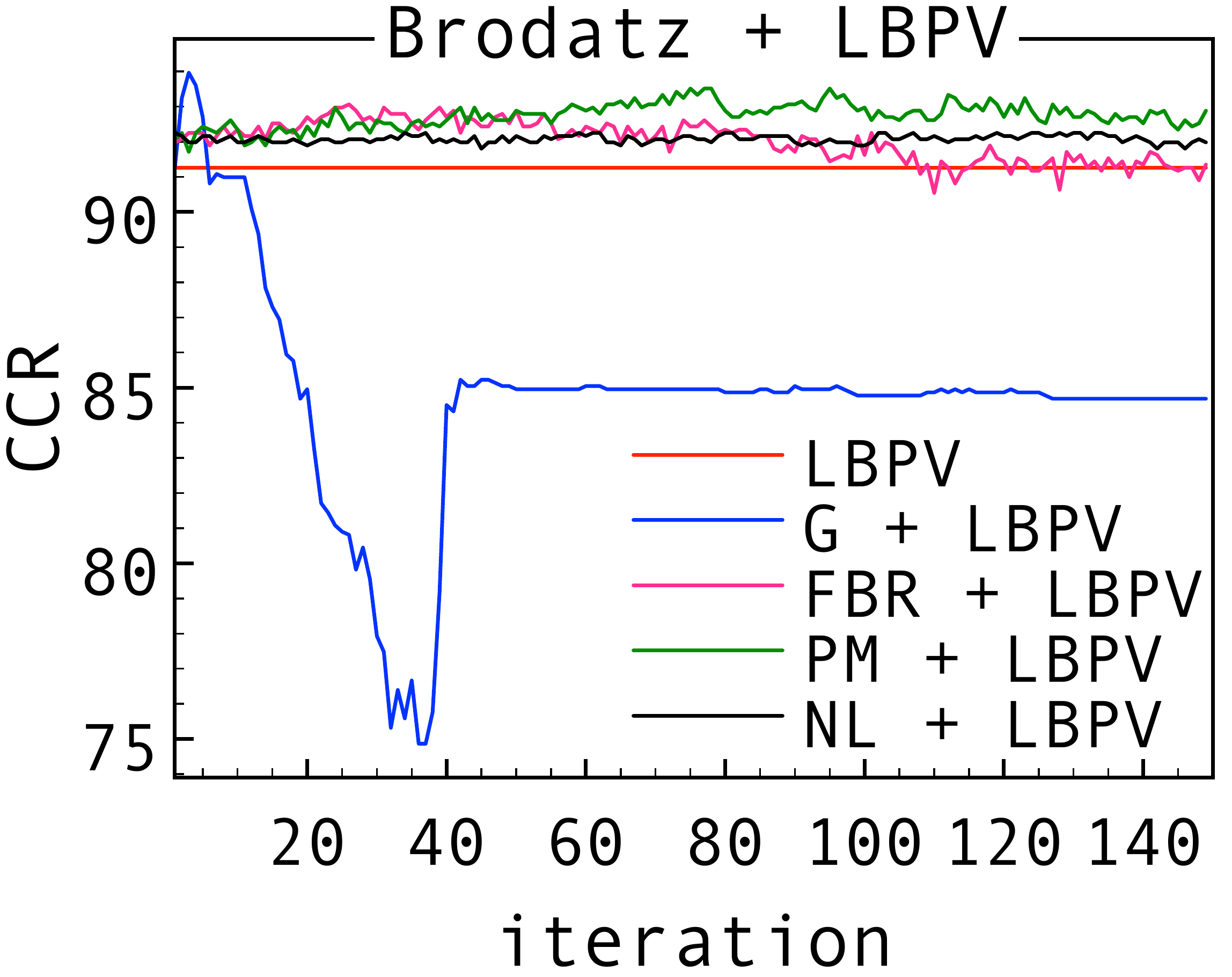} }}\subfloat[]{{\includegraphics[width=0.33\textwidth]{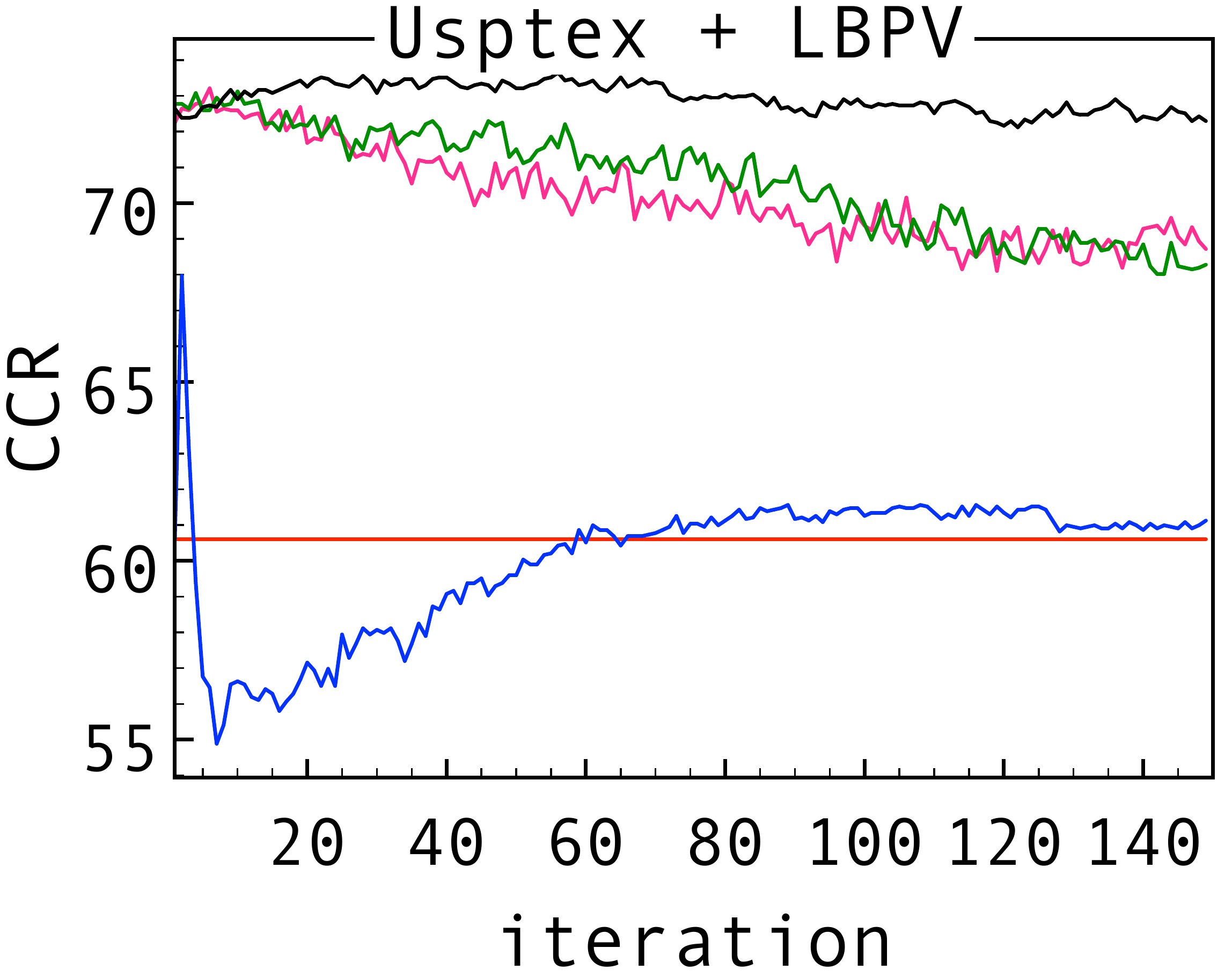} }} \subfloat[]{{\includegraphics[width=0.33\textwidth]{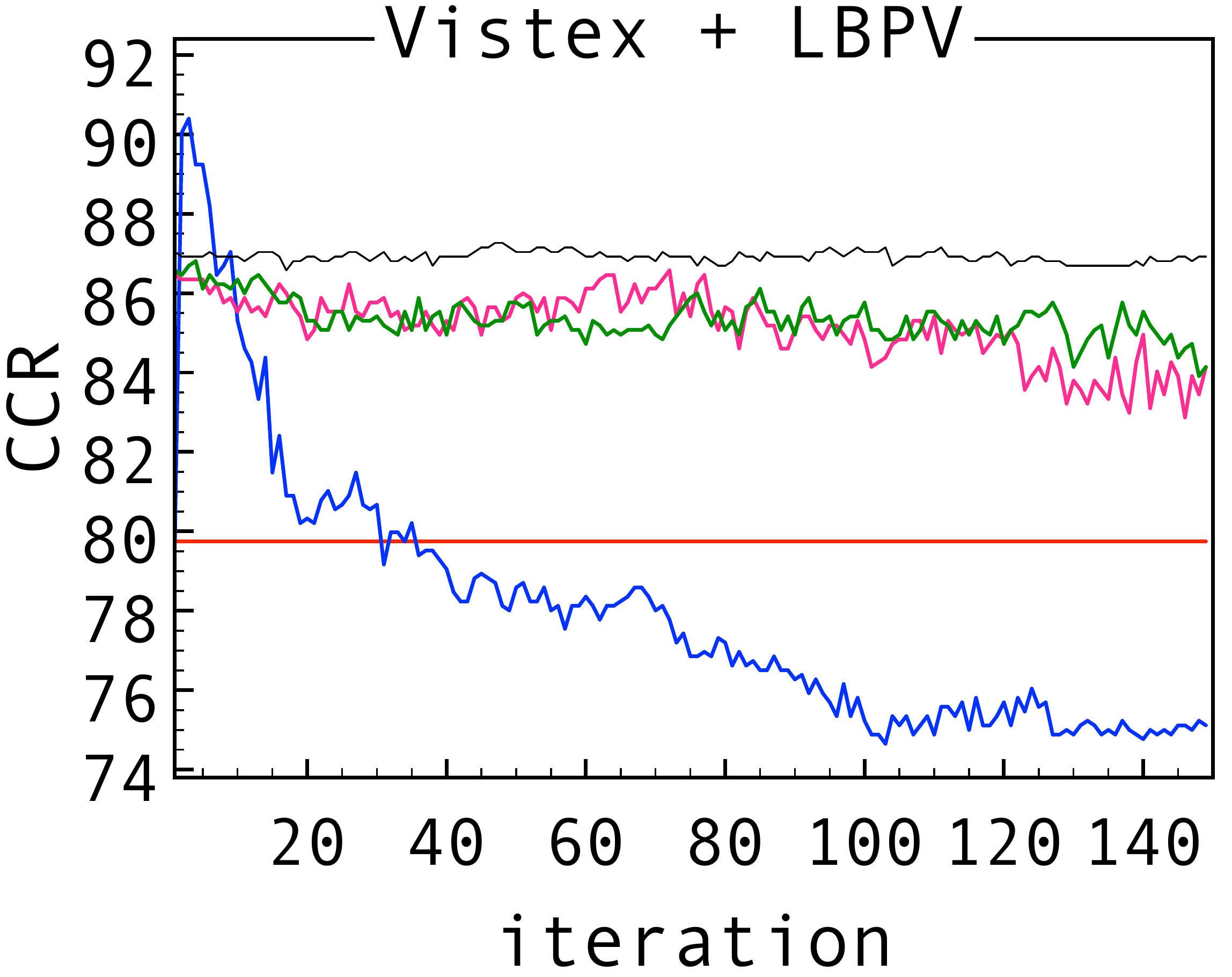} }}      
   \caption{Results of each iteration when preprocessing methods are combined with LBPV (Naive Bayes) } \label{plot:lbpv}
\end{figure}

\begin{figure}%
 \centering
 \subfloat[Brodatz]{{\includegraphics[width=0.33\textwidth]{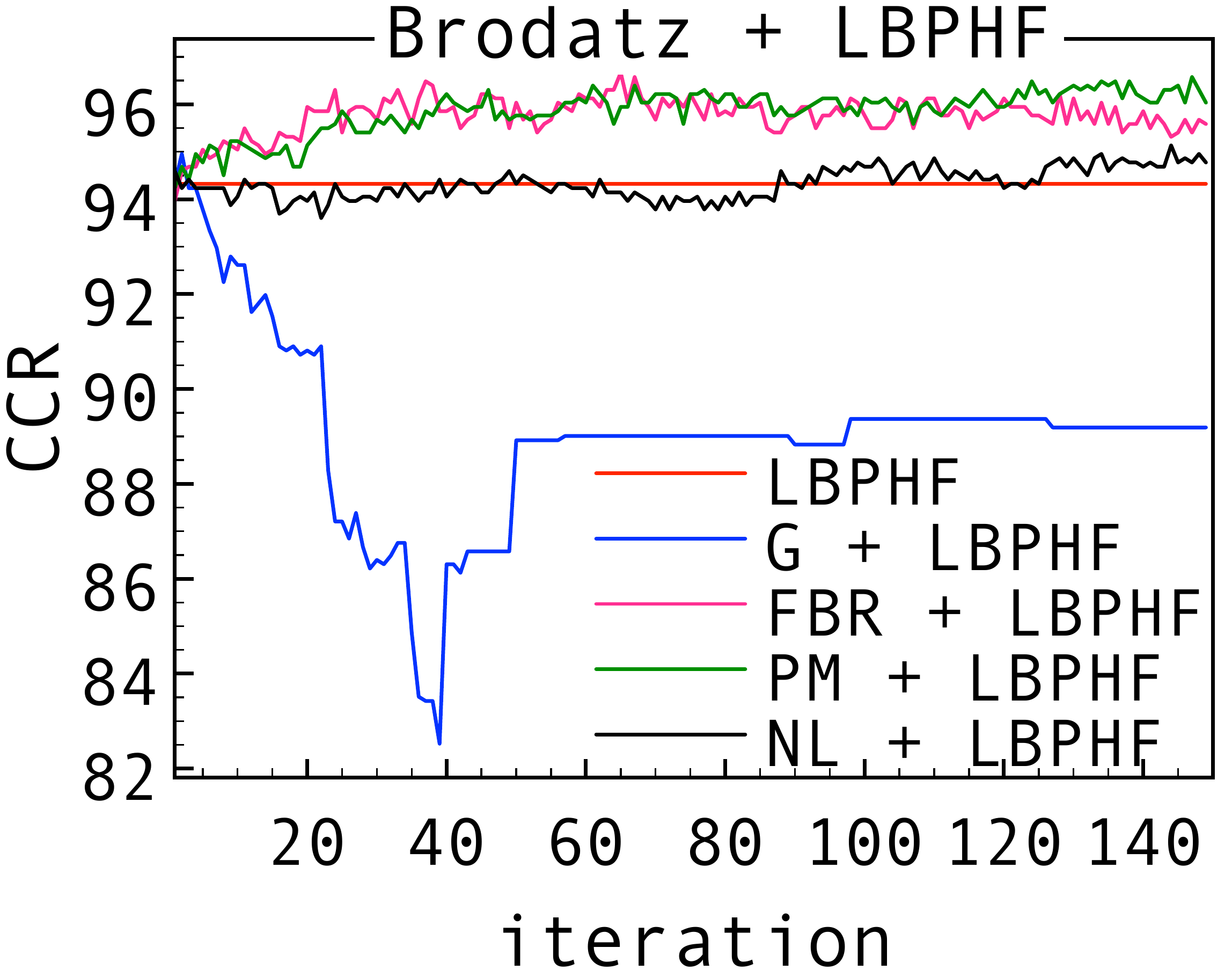} }}\subfloat[Usptex]{{\includegraphics[width=0.33\textwidth]{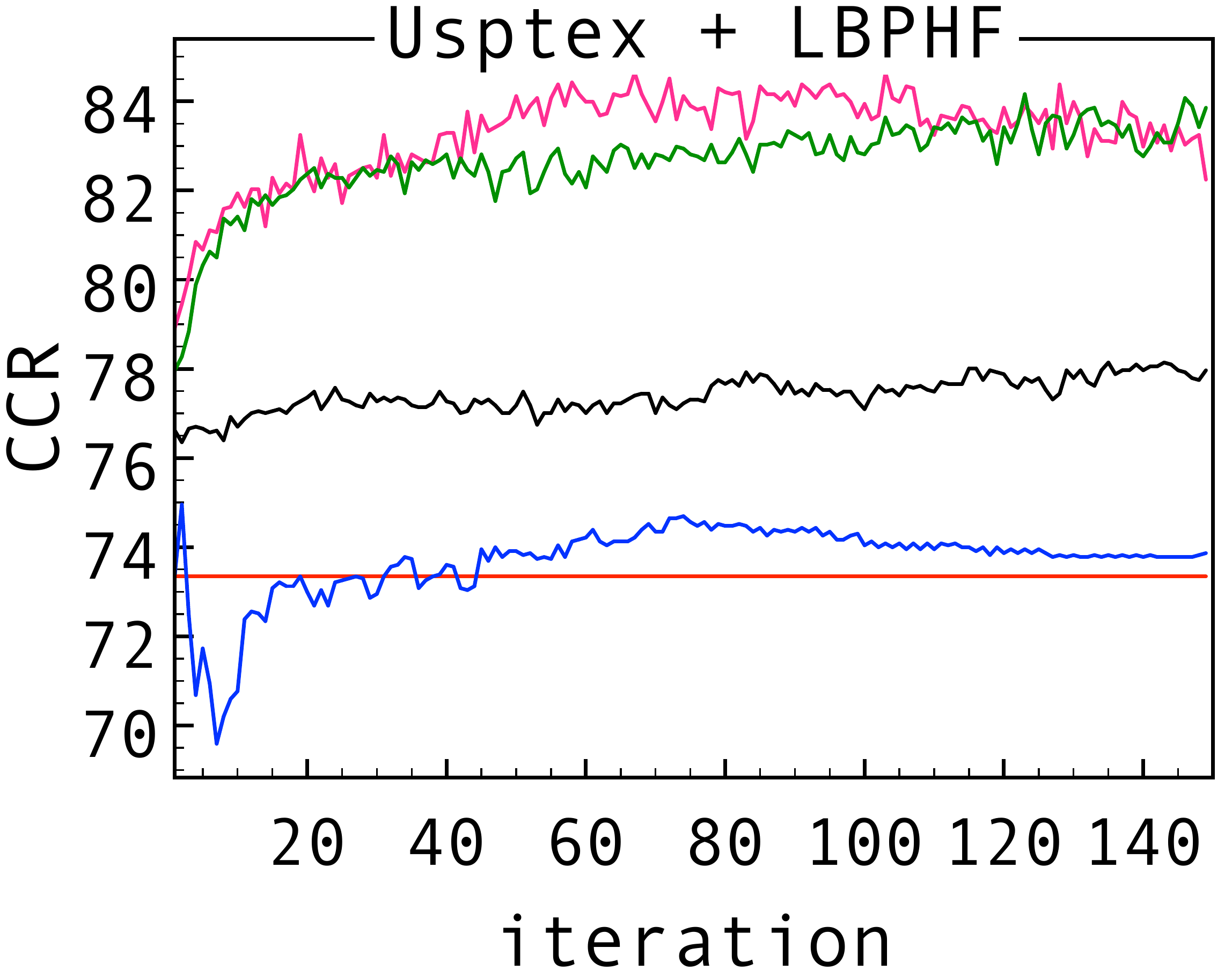} }}\subfloat[Vistex]{{\includegraphics[width=0.33\textwidth]{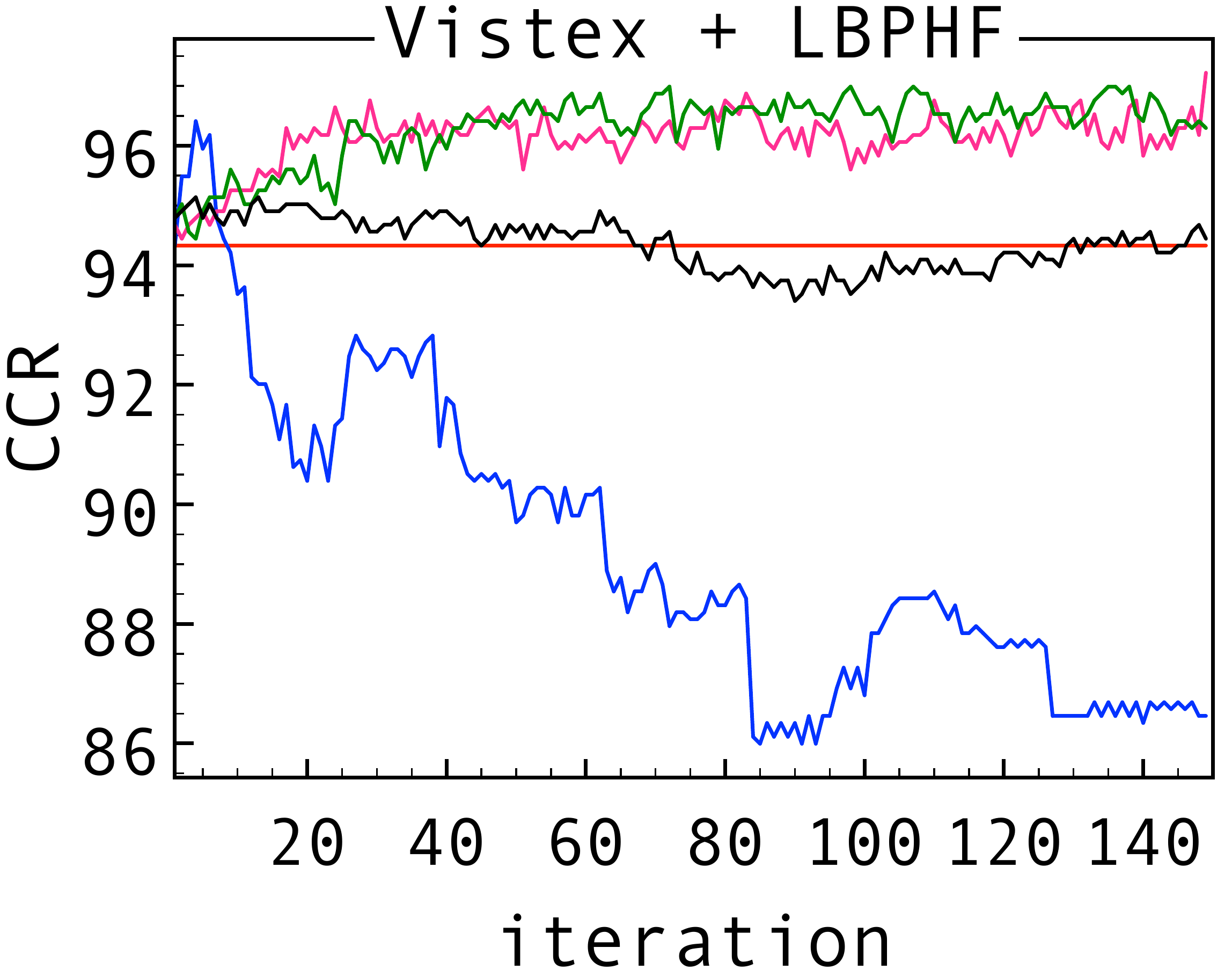} }}%
    \caption{Results of each iteration when preprocessing methods are combined with LBPHF (Naive Bayes) }\label{plot:lbphf}
  \end{figure}

\begin{figure}%
  \centering
 \subfloat[Brodatz]{{\includegraphics[width=0.33\textwidth]{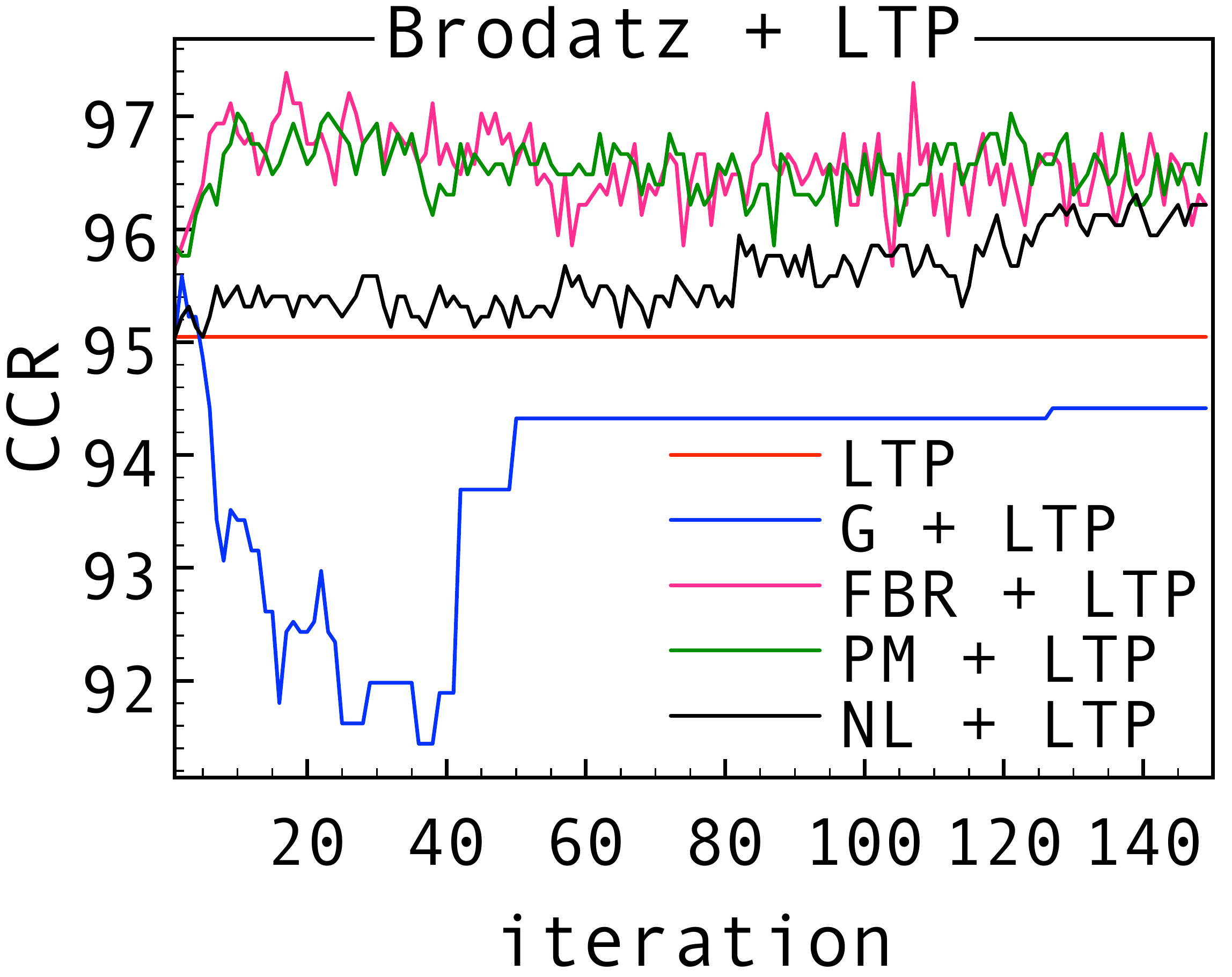} }}\subfloat[Usptex]{{\includegraphics[width=0.33\textwidth]{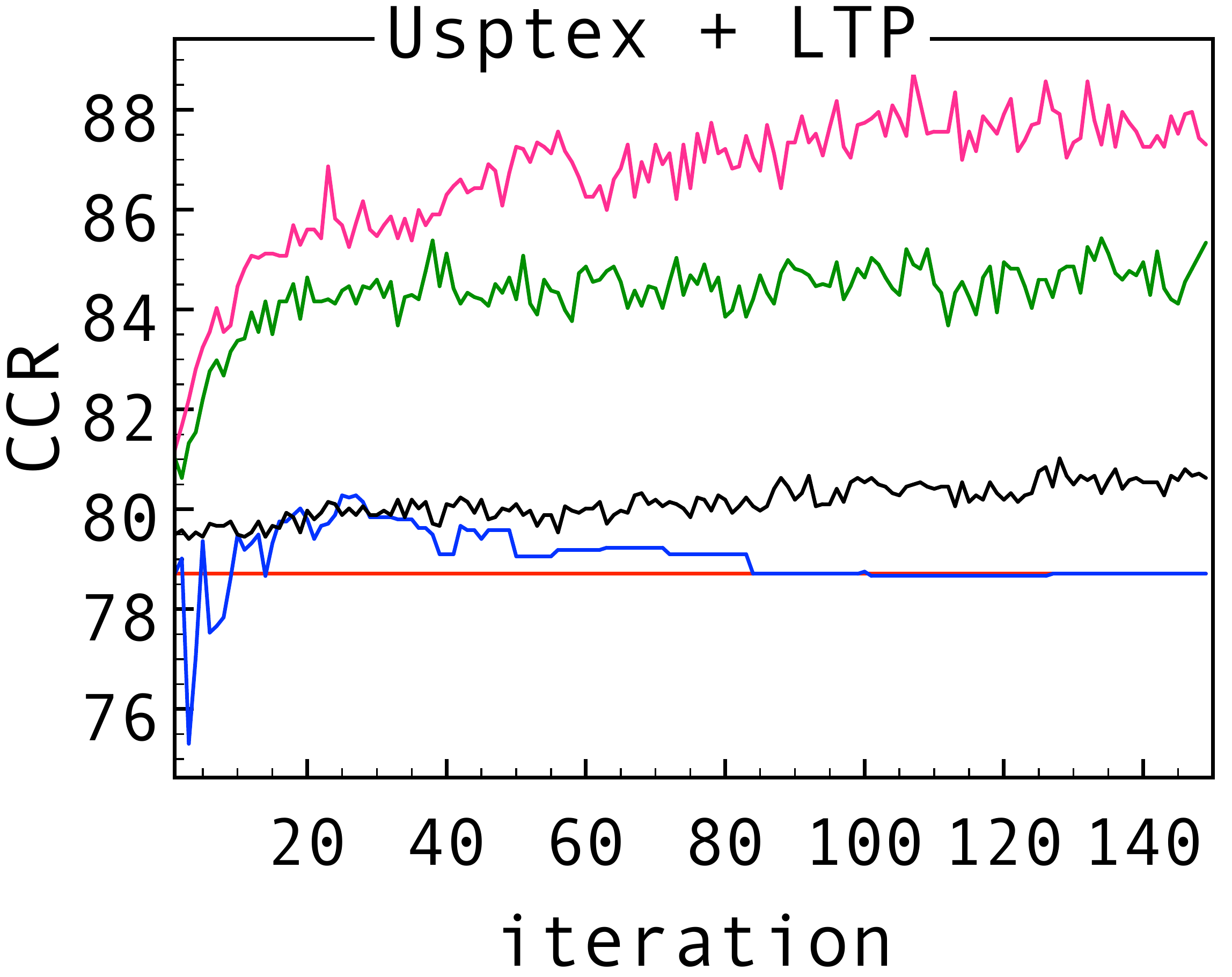} }}\subfloat[Vistex]{{\includegraphics[width=0.33\textwidth]{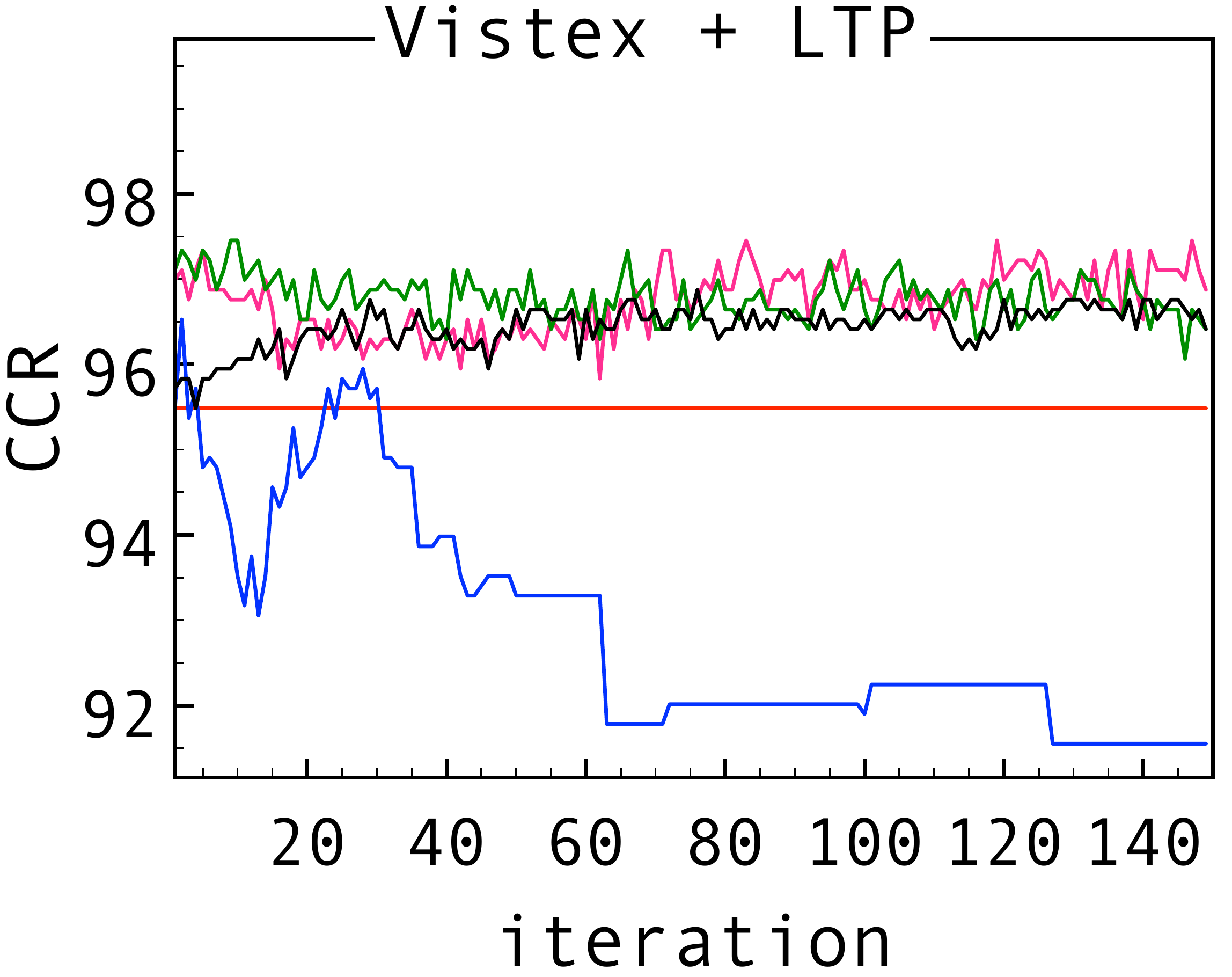}}}%
  \caption{Results of each iteration when preprocessing methods are combined with LTP (Naive Bayes)} \label{plot:ltp}
\end{figure}

\begin{figure}%
  \centering
 \subfloat[Brodatz]{{\includegraphics[width=0.33\textwidth]{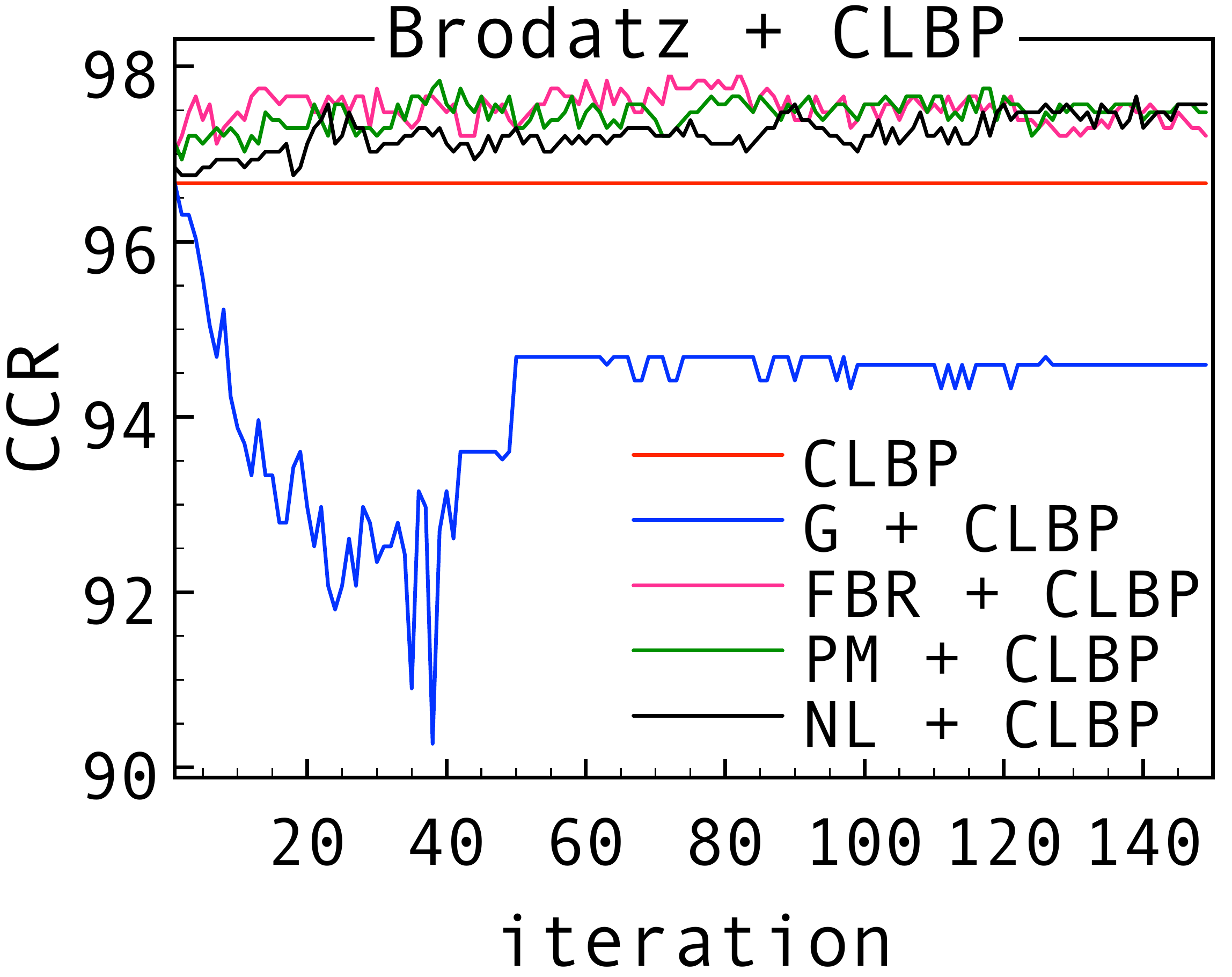} }} \subfloat[Usptex]{{\includegraphics[width=0.33\textwidth]{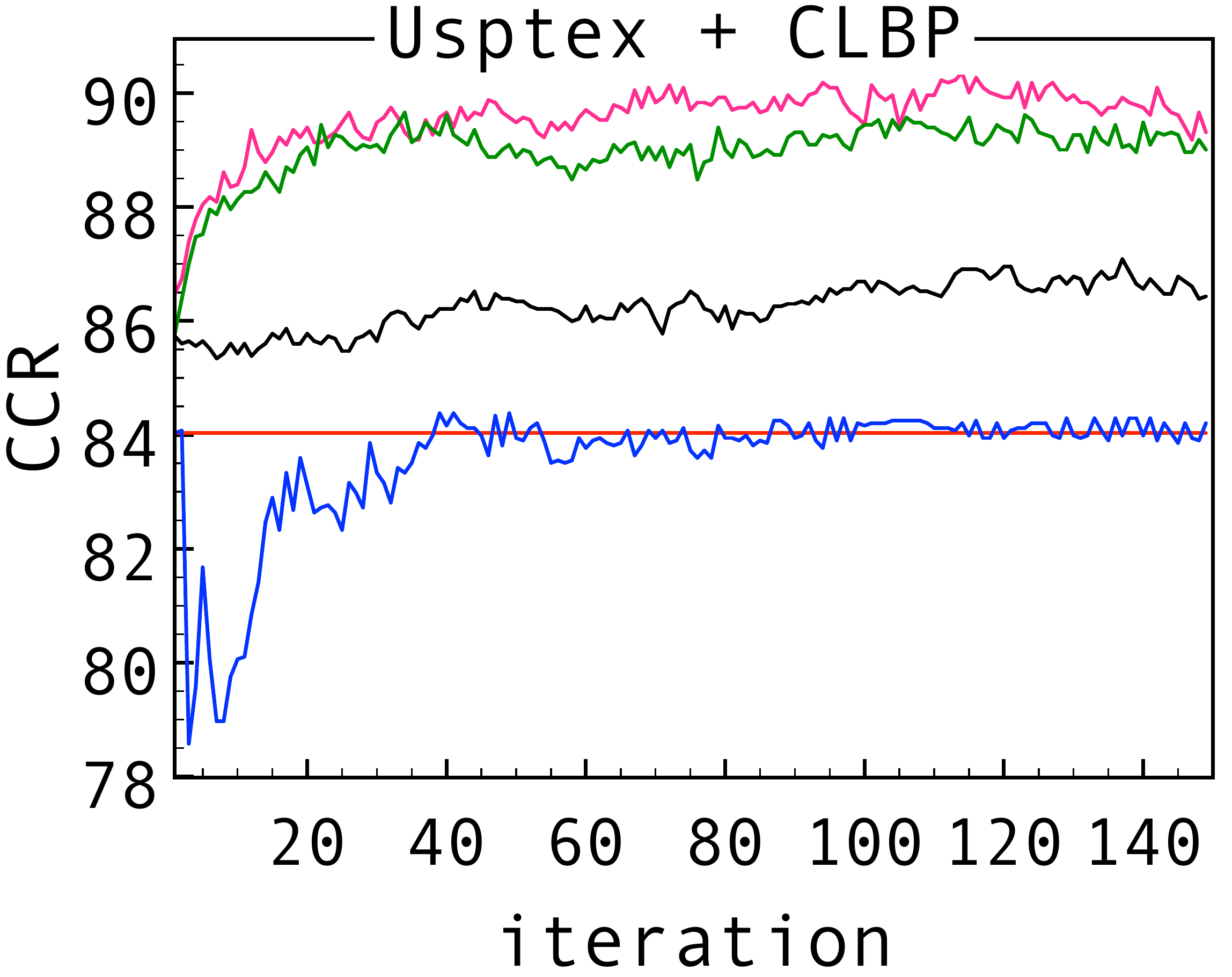} }}\subfloat[Vistex]{{\includegraphics[width=0.33\textwidth]{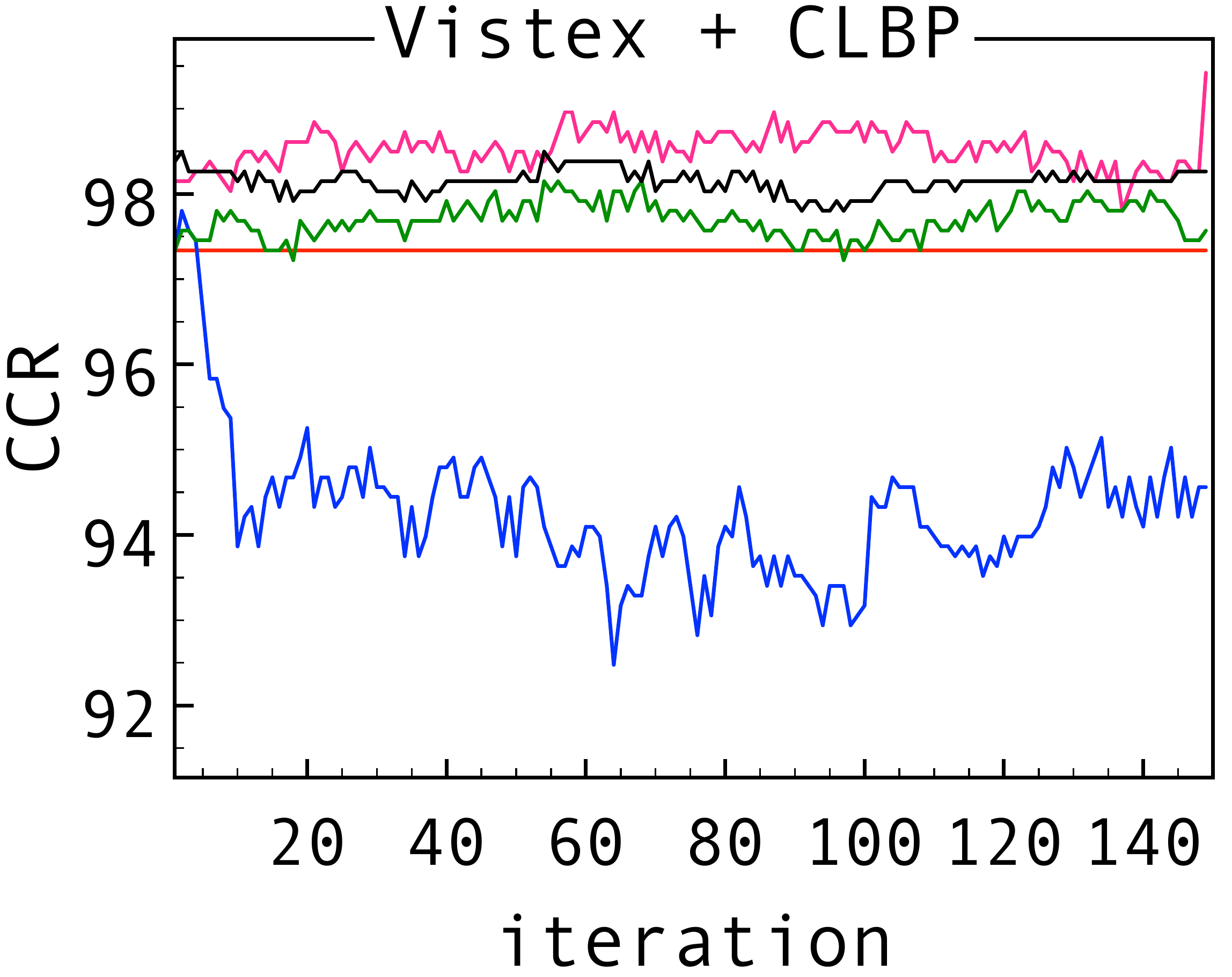} }}%
  \caption{Results of each iteration when preprocessing methods are combined with CLBP (Naive Bayes)} \label{plot:clbp}
\end{figure}

Images \ref{plot:lbp}, \ref{plot:clbp}, \ref{plot:lbpv}, \ref{plot:lbphf},  \ref{plot:ltp} and \ref{plot:cslbp} show that, while the use of a Gaussian kernel, the classical preprocessing approach, does result in an improvement, in general, the improvement is less pronounced than with non-isotropic diffusion methods. Also, the improvements are usually perceived only for small scales due to the fact that edges disappear at higher scales, an inherent property of the method. In the case of LBP, LBPV, LTP and CSLBP, all 150 iterations of anisotropic image processing improve texture extraction and increase the recognition rate. However, for LBPHF some iterations output a negative result for Brodatz and Vistex. In addition, graphs show a high similarity among preprocessing methods PM and FBR. However, as stated in \cite{guidotti2014anisotropic}, the classic anisotropic diffusion results in images with evident staircasing effect and, eventually, to the lower recognition rates. 


%
\begin{table}[hp!]
\centering
\caption{Best results for KNN (k = 1) of all combinations considering the 150 iterations}
\label{table:knn}
\resizebox{\textwidth}{!}{%
\begin{tabular}{|c|ll|ll|ll|}
\hline
\multirow{3}{*}{Method}         & \multicolumn{6}{c|}{KNN, k = 1}                                                               \\ \cline{2-7}  & \multicolumn{2}{c|}{Brodatz}                   & \multicolumn{2}{c|}{Vistex}               & \multicolumn{2}{c|}{Usptex} \\
 &      &  best i
 &      &  best i 
 &      & best i          \\ \hline
 
LBP&93.96(1.70)& - &
94.21 (2.74)& - &
73.65 (2.47)& - \\
G+LBP&95.59 (1.19)&4&
97.69 (1.90)&10&
79.01 (1.58)&5 \\
PM+LBP&\textbf{97.66 }(0.77)&22&
98.61 (1.75)&73&
82.72 (1.69)&31 \\
FBR+LBP&97.21 (0.83)&20&
\textbf{98.73} (1.36)&82&
\textbf{84.55} (2.29)&33 \\
NL+LBP&96.76 (1.77)&124&
95.60 (2.21)&130&
77.84 (1.57)&140 \\ \hline

CLBP&97.21(0.99)& - &
98.96 (0.98)& - &
83.16 (2.10)& - \\ 
G+CLBP&97.21 (0.99)&2&
98.96 (1.43)&2&
85.21 (2.59)&1\\
PM+CLBP&97.84 (0.97)&134&
99.54 (0.98)&14&
88.57 (1.81)&127\\
FBR+CLBP&\textbf{97.93} (1.06)&48&
\textbf{99.88} (0.78)&149&
\textbf{89.44} (1.39)&53\\ 
NL+CLBP&97.84 (0.77)&50&
99.42 (1.46)&10&
85.86 (2.36)&145\\ \hline

\hline

\hline

\hline

LBPV&94.23(1.76)& - &
92.82 (3.14)& - &
73.30 (3.62)& - \\
G+LBPV&95.41 (1.50)&4&
\textbf{94.56 }(3.16)&3&
75.57 (2.62)&10 \\
PM+LBPV&95.68 (1.06)&43&
94.44 (4.59)&37&
79.14 (2.66)&10 \\
FBR+LBPV&\textbf{96.22 }(1.44)&64&
\textbf{94.56} (4.01)&22&
\textbf{79.71} (2.65)&36 \\
NL+LBPV&95.23 (1.31)&105&
93.75 (3.95)&60&
79.06 (3.00)&107 \\ \hline

LBPHF&93.33(1.86)& - &
82.29 (3.41)& - &
56.11 (1.50)& - \\
G+LBPHF&94.95 (1.14)&6&
90.39 (2.54)&10&
66.36 (2.11)&5\\
PM+LBPHF&96.31 (0.79)&38&
92.82 (3.37)&138&
69.42 (2.99)&24\\ 
FBR+LBPHF&\textbf{96.40 }(1.42)&13&
\textbf{93.52} (2.19)&53&
\textbf{74.21} (2.18)&110\\
NL+LBPHF&94.95 (1.52)&148&
87.27 (1.99)&150&
64.83 (2.90)&120\\ \hline

LTP&93.96(1.70)& - &
94.21 (2.74)& - &
73.65 (2.47)& - \\
G+LTP&95.59 (1.19)&4&
97.69 (1.90)&10&
79.01 (1.58)&5\\
PM+LTP&\textbf{97.66} (0.77)&22&
98.61 (1.75)&73&
82.72 (1.69)&31\\
FBR+LTP&96.94 (1.37)&8&
\textbf{98.73} (1.36)&82&
\textbf{84.55} (2.29)&33\\
NL+LTP&96.76 (1.77)&124&
95.60 (2.21)&130&
77.84 (1.57)&140\\ \hline

CSLBP&86.49(2.33)& - &
82.87 (4.76)& - &
56.50 (3.59)& - \\
G+CSLBP&91.89 (2.13)&4&
\textbf{93.63} (2.51)&7&
\textbf{69.46 }(2.09)&6\\
PM+CSLBP&\textbf{92.43} (1.27)&95&
91.44 (3.00)&93&
68.98 (2.90)&140\\
FBR+CSLBP&\textbf{92.43 }(1.35)&65&
92.82 (3.10)&62&
67.84 (2.77)&44\\
NL+CSLBP&90.63 (1.91)&142&
86.00 (4.74)&119&
62.52 (2.41)&147\\ \hline

\end{tabular}%
}
\end{table}

\begin{table}[hp!]
\centering
\caption{Best results for Naive Bayes of all combinations considering the 150 iterations}
\label{table:naivebayes}
\resizebox{\textwidth}{!}{%
\begin{tabular}{|c|ll|ll|ll|}
\hline
\multirow{3}{*}{Method}         & \multicolumn{6}{c|}{Naive Bayes}                                                                                      \\ \cline{2-7} 
                                & \multicolumn{2}{c|}{Brodatz}   & \multicolumn{2}{c|}{Vistex}   & \multicolumn{2}{c|}{Usptex} \\
                                &                      & best i &                          & best i &           & best i          \\ \hline
                                
LBP&95.05(1.60)& - &
95.49 (3.27)& - &
78.71 (1.44)& - \\
G+LBP&95.50 (1.82)&3&
96.53 (1.32)&3&
81.33 (2.12)&1 \\
PM+LBP&96.94 (1.20)&11&
\textbf{97.45} (1.82)&10&
85.47 (1.63)&39 \\
FBR+LBP&\textbf{97.57} (1.80)&18&
\textbf{97.45} (1.20)&84&
\textbf{88.53} (2.47)&108 \\
NL+LBP&96.31 (1.63)&140&
96.88 (3.08)&77&
81.02 (1.47)&129 \\ \hline




LBPV&91.26(2.72)& - &
79.75 (4.33)& - &
60.60 (1.76)& - \\
G+LBPV&\textbf{93.96} (2.34)&4&
\textbf{90.39} (3.02)&4&
68.41 (3.04)&1 \\
PM+LBPV&93.51 (2.76)&76&
86.81 (4.10)&1&
73.12 (2.87)&11 \\
FBR+LBPV&93.06 (2.55)&27&
86.92 (3.98)&1&
73.21 (2.73)&7 \\
NL+LBPV&92.25 (2.86)&2&
87.27 (3.98)&48&
\textbf{73.65 }(2.89)&57 \\ \hline

CLBP&96.67(1.35)& - &
97.34 (1.50)& - &
84.03 (2.17)& - \\
G+CLBP&96.85 (1.48)&1&
97.80 (1.06)&3&
85.86 (2.14)&1\\
PM+CLBP&97.84 (1.20)&40&
98.15 (1.74)&55&
89.66 (2.31)&35\\ 
FBR+CLBP&\textbf{97.93} (1.14)&73&
\textbf{99.42 }(1.35)&150&
\textbf{90.36} (2.24)&115\\ 
NL+CLBP&97.66 (0.87)&140&
98.50 (2.46)&3&
87.09 (2.35)&138\\
\hline

LBPHF&94.32(1.65)& - &
94.33 (2.09)& - &
73.34 (2.33)& - \\ 
G+LBPHF&94.95 (1.46)&3&
96.41 (1.90)&5&
79.41 (2.19)&1\\
PM+LBPHF&96.58 (1.41)&148&
96.99 (2.39)&73&
84.16 (1.65)&124\\
FBR+LBPHF&\textbf{96.67} (2.08)&66&
\textbf{97.22} (1.57)&150&
\textbf{84.64} (2.27)&68\\
NL+LBPHF&95.14 (1.41)&145&
95.14 (1.81)&5&
78.14 (2.34)&136\\
\hline

LTP&95.05(1.60)& - &
95.49 (3.27)& - &
78.71 (1.44)& - \\
G+LTP&95.59 (1.77)&3&
96.53 (1.28)&3&
81.28 (2.02)&1\\
PM+LTP&97.03 (1.20)&11&
\textbf{97.45} (1.82)&10&
85.43 (2.23)&135\\
FBR+LTP&\textbf{97.39} (1.66)&18&
\textbf{97.45} (1.20)&84&
\textbf{88.74} (2.39)&108\\
NL+LTP&96.31 (1.63)&140&
96.88 (3.08)&77&
81.02 (1.29)&129\\ \hline

CSLBP&86.40(2.27)& - &
81.37 (3.45)& - &
58.16 (2.82)& - \\
G+CSLBP&90.45 (1.99)&5&
91.09 (3.14)&6&
68.24 (3.18)&6\\
PM+CSLBP&91.89 (1.80)&130&
90.16 (2.71)&148&
74.83 (1.83)&146\\
FBR+CSLBP&\textbf{92.07 }(2.17)&83&
\textbf{91.55} (2.55)&86&
\textbf{76.18} (2.55)&142\\ 
NL+CSLBP&88.92 (1.69)&134&
85.19 (3.33)&146&
64.05 (3.15)&122\\\hline

\end{tabular}%
}
\end{table}
\section{Conclusion} \label{ref:conclusion}

This paper evaluates the influence of different diffusion methods, three of which are nonlinear and anisotropic, and one of which is linear and isotropic, as a preprocessing tool generating extended feature vectors for known texture classifiers.  The goal was to verify whether the use of such a preprocessing step prior to feature extraction would enhance image characteristics and consequently increase texture recognition. Six texture analysis algorithms were tested (LBP, LBPV, CLBP, LBPHF, LTP and CSLBP) in combination with two different classifiers (KNN with $k=1$ and Naive Bayes). 

The experiments demonstrated that the use of any of these preprocessing methods did improve texture extraction and pattern recognition. In particular, the combination FBR + CLBP delivered the highest gain for all datasets. 

 LBP and its extended methods (cited in Section \ref{sec:fem} analyzes the contrast of texture and the use of anisotropic diffusion proved helpful in improving this features while removing unecessary artifacts, which is very interesting for pattern recognition. As for the question of which methods to use in practice, Tables \ref{table:knn} and \ref{table:naivebayes} help in identifying a good candidate for the feature extraction methods considered in this paper. More in general, however, if one chooses a feature extraction not considered here, the tip would be to first apply linear isotropic diffusion with small number of scales and check the results. 
 
\section*{Acknowledgments}
 
The authors acknowledges support from CNPq (National Counsel of Technological and Scientific Development) (Grant Nos. 132409/2014-3, 307797/2014-7 and 484312/2013-8), FAPESP (Grant No. 14/08026-1) and CAPES (Coordination for the Improvement of Higher Education Personnel) (Grant No. 9056169/D).


\end{document}